\documentclass[sigconf,nonacm]{acmart}
\usepackage{times}  
\usepackage{helvet}  
\usepackage{booktabs}
\usepackage{courier}  
\usepackage{hyperref}

\usepackage{graphicx} 

\usepackage{natbib}  
\usepackage[labelfont=bf,font=small]{caption}
\usepackage{subcaption}
\usepackage{todonotes}
\DeclareCaptionStyle{ruled}{labelfont=normalfont,labelsep=colon,strut=off} 
\usepackage{algorithm}
\usepackage{algorithmic}

\usepackage{mathtools}
\usepackage{etoolbox}

\usepackage{newfloat}
\usepackage{listings}
\floatstyle{ruled}
\newfloat{listing}{tb}{lst}{}
\floatname{listing}{Listing}

\begin{document}  





\title{Learning Controllable 3D Level Generators}

\author{Zehua Jiang}
\affiliation{%
  \institution{NYU Tandon}
  \city{New York}
  \country{USA}}
\email{zehua.jiang@nyu.edu}

\author{Sam Earle}
\affiliation{%
  \institution{NYU Tandon}
  \city{New York}
  \country{USA}}
\email{sam.earle@nyu.edu}

\author{Michael Cerny Green}
\affiliation{%
  \institution{NYU Tandon}
  \city{New York}
  \country{USA}}
\email{mike.green@nyu.edu}

\author{Julian Togelius}
\affiliation{%
  \institution{NYU Tandon}
  \city{New York}
  \country{USA}}
\email{julian@togelius.com}


\begin{abstract}

Procedural Content Generation via Reinforcement Learning (\mbox{PCGRL}) foregoes the need for large human-authored data-sets and allows agents to train explicitly on functional constraints, using computable, user-defined measures of quality instead of target output.
We explore the application of PCGRL to 3D domains, in which content-generation tasks naturally have greater complexity and potential pertinence to real-world applications. Here, we introduce several PCGRL tasks for the 3D domain, Minecraft. These tasks will challenge RL-based generators using affordances often found in 3D environments, such as jumping, multiple dimensional movement, and gravity.
We train agents to optimize each of these tasks to explore the capabilities of existing in PCGRL.
The agents are able to generate relatively complex and diverse levels, and generalize to random initial states and control targets. Controllability tests in the presented tasks demonstrate their utility to analyze success and failure for 3D generators. We argue that these generators could serve both as co-creative tools for game designers, and as pre-trained environment generators in curriculum learning for player agents.

\end{abstract}

\begin{teaserfigure}
    \centering
    \includegraphics[trim={50 50 50 50},clip,width=0.19\linewidth]{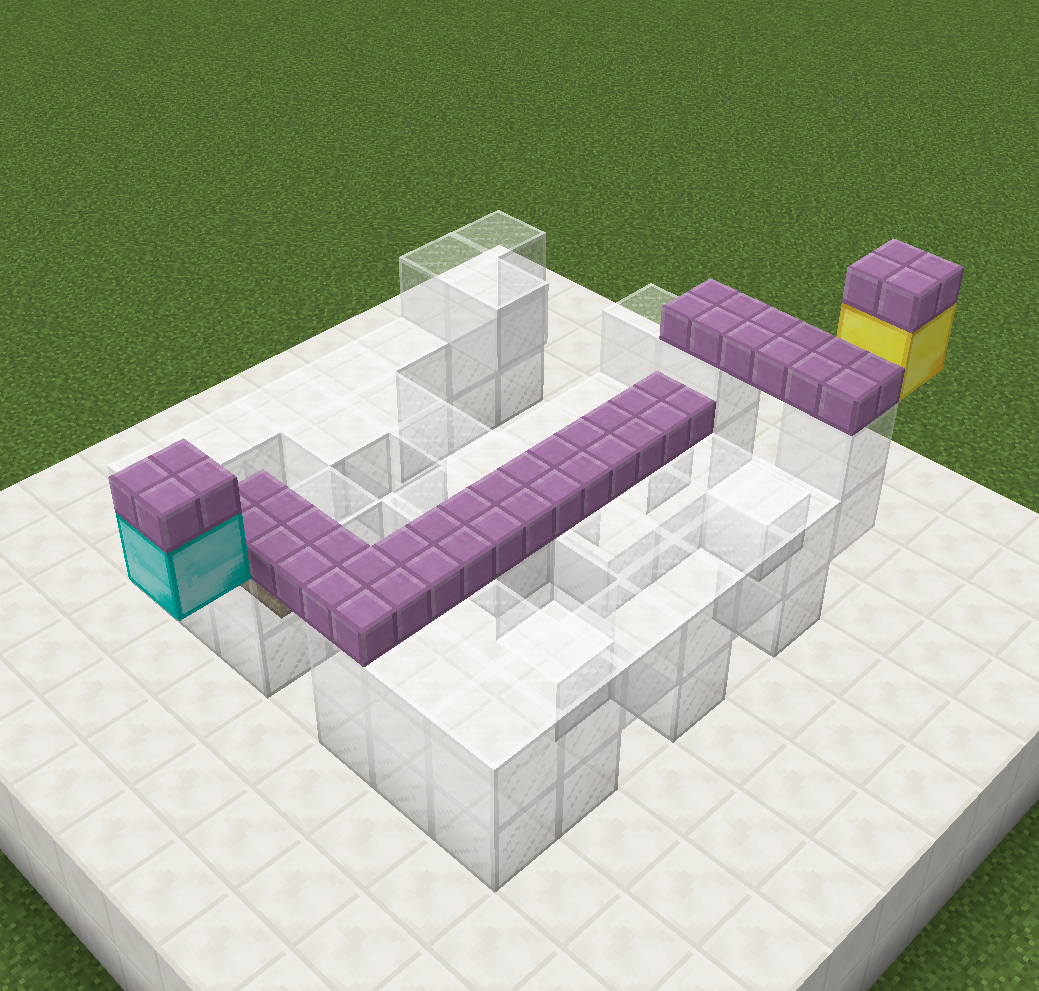}
    \includegraphics[trim={50 50 50 50},clip,width=0.19\linewidth]{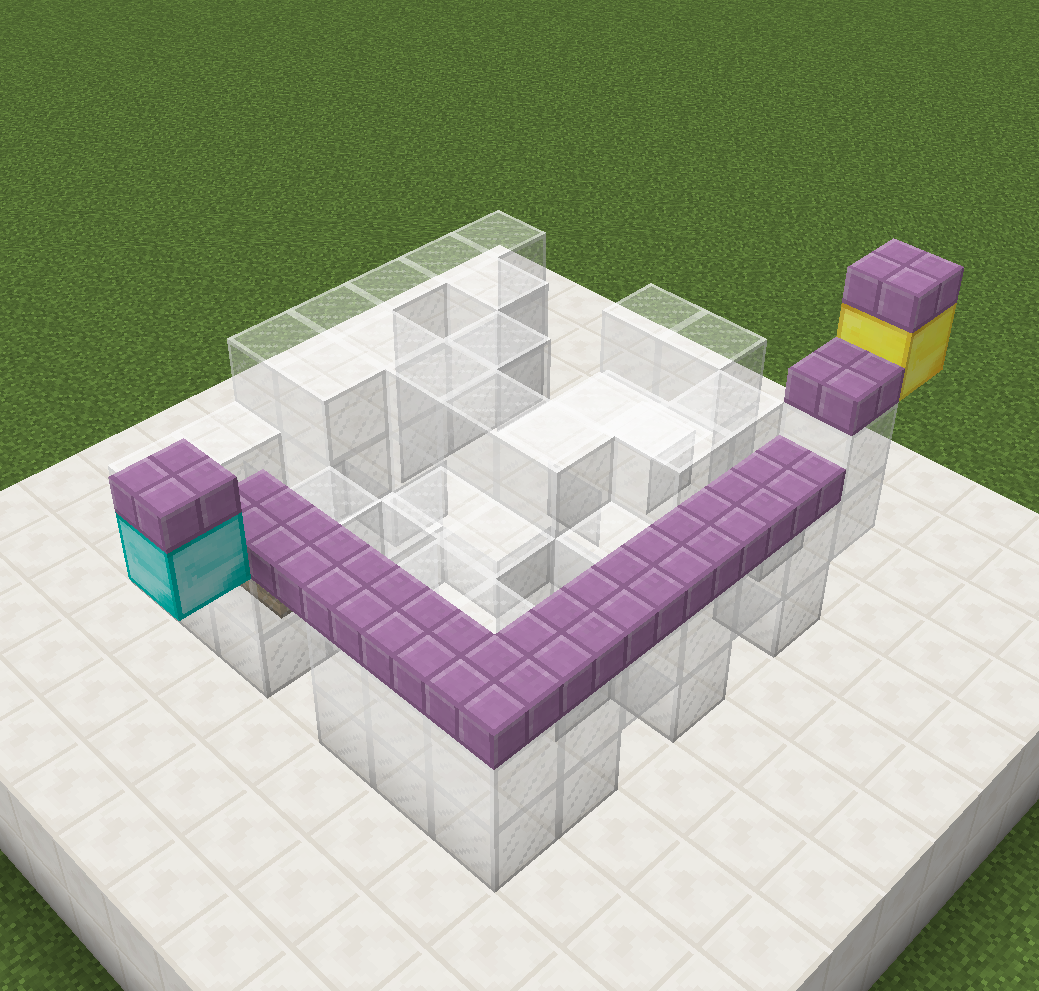}
    \includegraphics[trim={50 50 50 50},clip,width=0.19\linewidth]{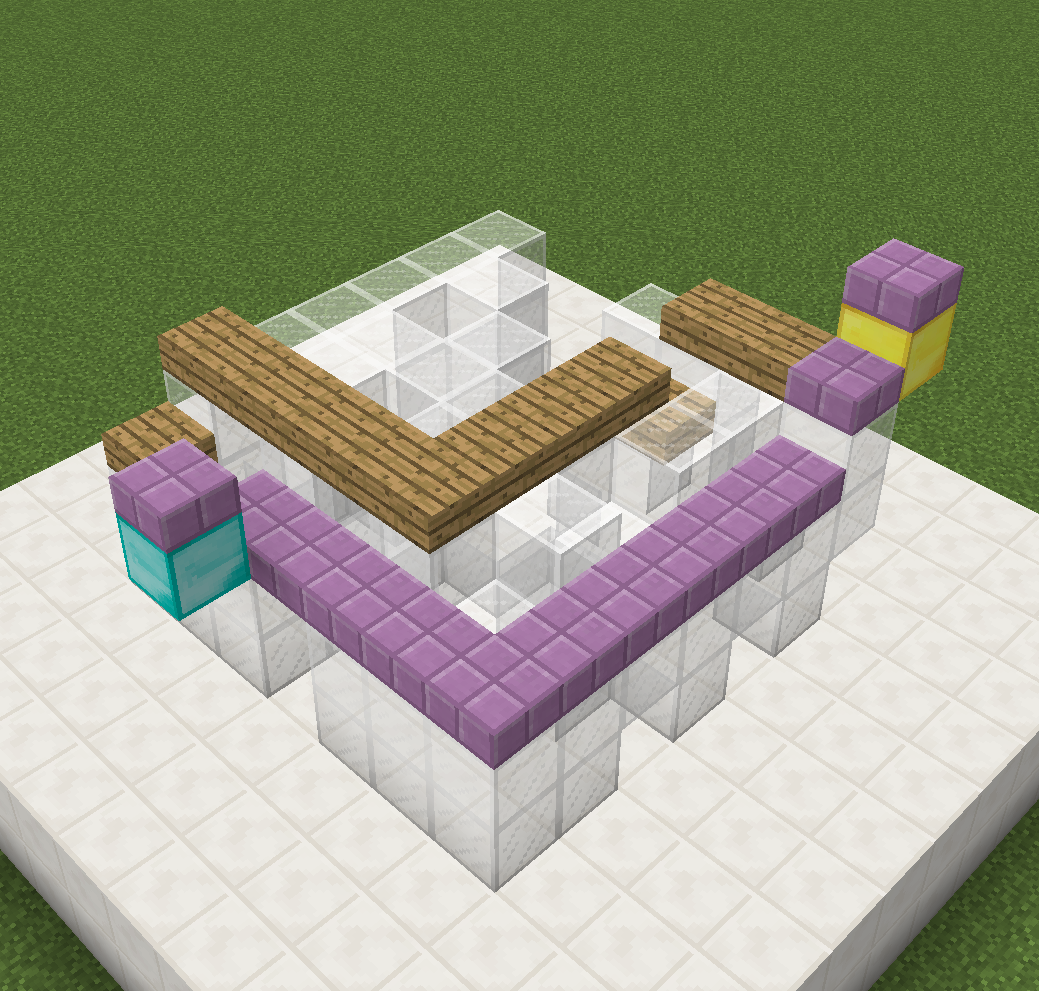}
    \includegraphics[trim={50 50 50 50},clip,width=0.19\linewidth]{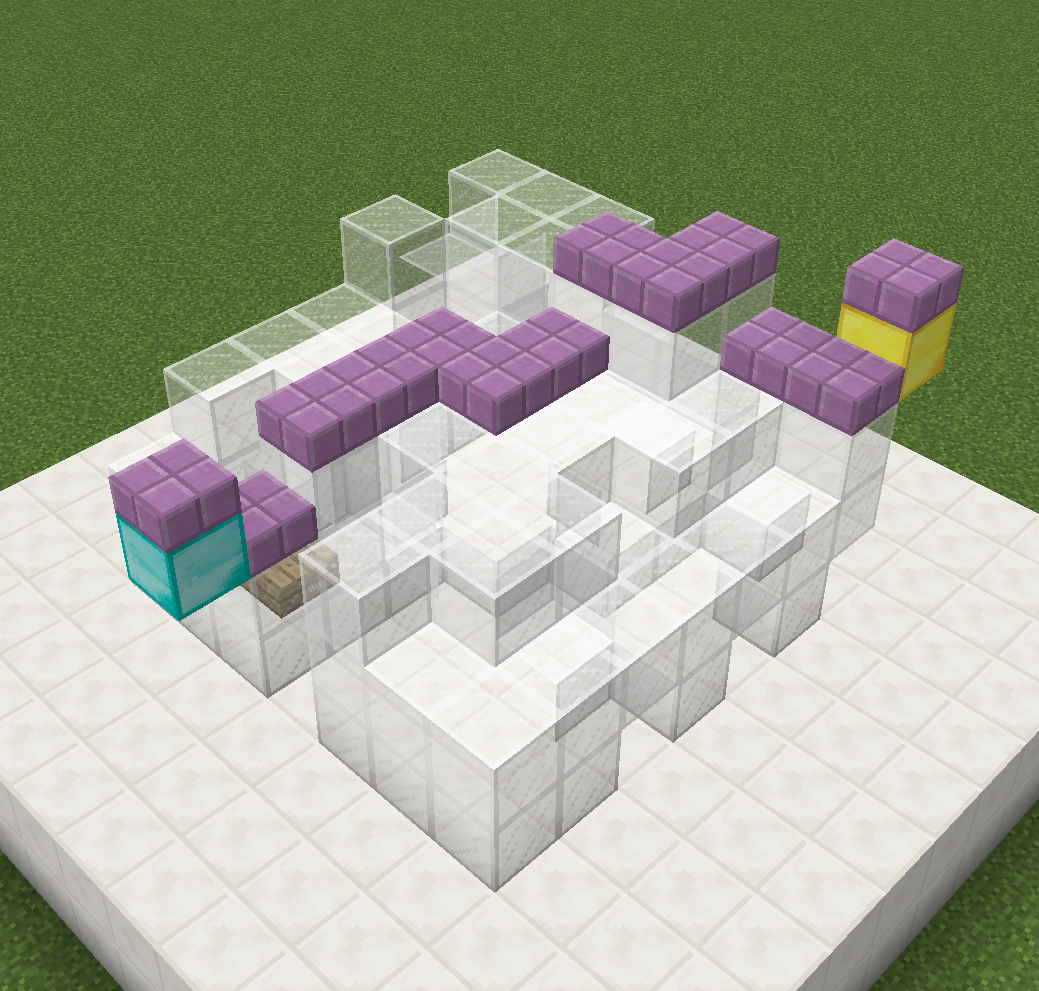}
    \includegraphics[trim={50 50 50 50},clip,width=0.19\linewidth]{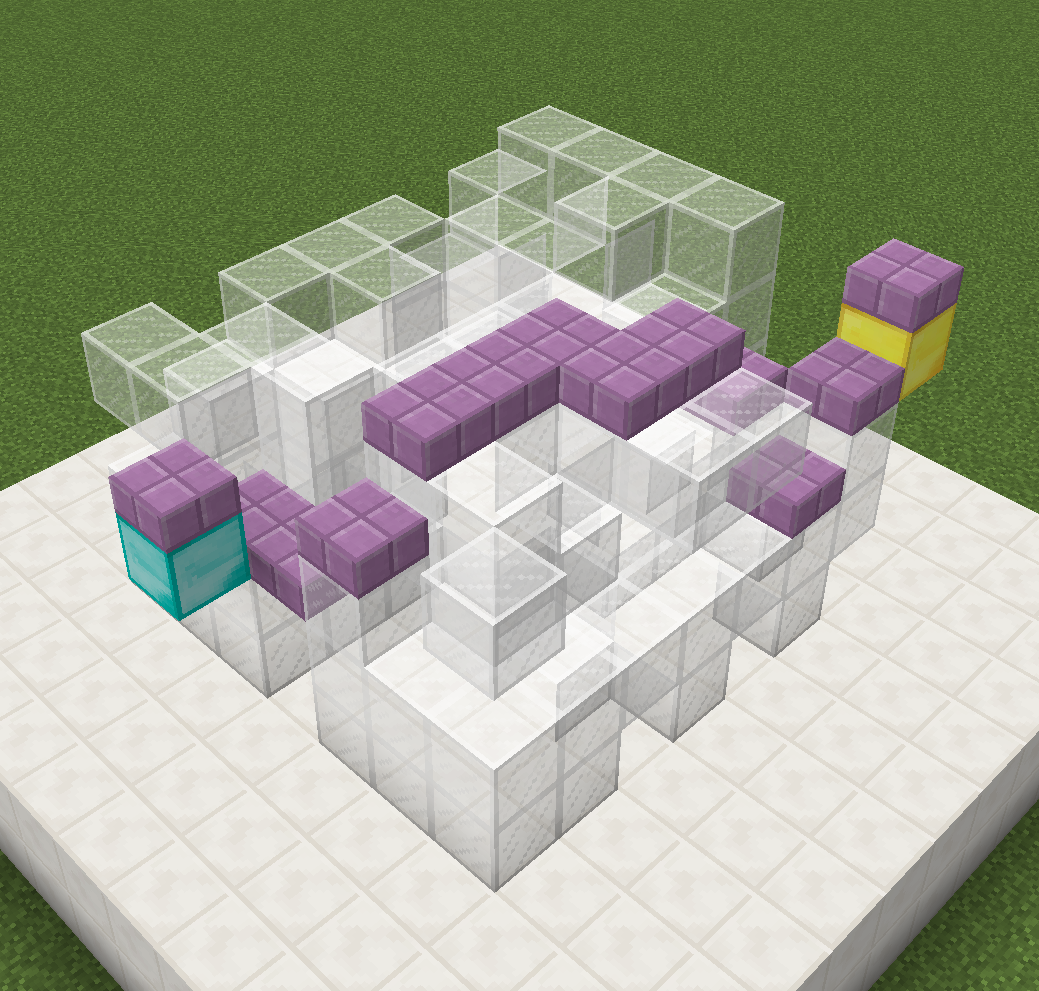}
    \caption{A generator connects two doors which are randomly placed on the $7 \times 7 \times 7$ border. The generator gradually complexifies the connected path with a staircase-like structure, while ensuring it is traversable for the player. To make the reward for training more dense, we also encourage the agent to expand the longest path from the entrance, even if it may not connect to the exit. Here, we show how the agent complicates the path between doors, starting in an empty $7 \times 7 \times 7$ space. We use the purple slab stone to denote the connected path and wooden slab to denote the longest path from the entrance.
    }
    \label{fig:door_episode}
\end{teaserfigure}
                     
\maketitle


\section{Introduction}
Most approaches applying machine learning to Procedural Content Generation (PCG) train models via supervised lerning on datasets of human-authored artifacts.
These models can often generate plausible-looking output, though there is no way of ensuring they have learnt to abide by the functional constraints of, e.g., game levels.

PCG via Reinforcement Learning (PCGRL) re-frames the problem of content-generation as an iterative task, in which the generator is rewarded at each step for editing the level to make it more playable, complex, or otherwise bringing it closer to satisfying some high-level user-defined metrics acting as a measure of quality. PCGRL thus foregoes the need for datasets while also explicitly training the generative model to abide by relevant functional constraints. We can contrast this against search-based PCG methods like evolution, which generates and tests content as an iterative process as well. However, this generate-and-test process can be prohibitively expensive because of the cost of the fitness function, especially when compared to the relatively cheap computational cost of RL inference (RL needs a reward function only during training).



Until now, PCGRL has been applied mostly to 2D, grid-based domains (such as mazes and dungeon crawlers), but in this paper we propose using a 3D environment. 3D content is interesting for multiple reasons. Most obviously, the real world is in three dimensions, at least on a macroscopic scale. This makes the ability to generate 3D content \emph{a must} for a method aiming to create anything that could be fabricated as a real-world object. Even when not fabricating things for the real world, many games and simulations model or are highly inspired by real-world scenarios in 3D.

3D games often incorporate gravity, which in turn poses new and interesting challenges for the design of levels and other functional content. For example, many paths become non-reversible, because an agent can fall down but not up. They also often feature first-person perspectives (unlike 2D games, which are often viewed in a third-person perspective) which poses challenges but also creates design affordances in terms of occlusion, visibility, and perspective. Beyond this, three-dimensional games are almost by necessity larger in terms of content (measured in the number of ``atomic'' content units).
 
In this work, we apply PCGRL to a 3D domain, the open-world first-person game \textit{Minecraft}, by training generators to produce complex mazes and dungeon-crawler/platformer levels.\footnote{Code is available at \href{https://github.com/smearle/control-pcgrl}{https://github.com/smearle/control-pcgrl}}
We introduce several tasks for 3D PCGRL, on which we train models to both demonstrate the effectiveness of a known PCGRL architecture~\cite{khalifa2020pcgrl} as well as showcase the utility of the 3D tasks for the advancement of PCGRL systems. We believe that these tasks showcase key affordances and challenges of 3D content and are therefore good evaluation tools for PCGRL models and training algorithms. 

\section{Related work}
Procedural Content Generation (PCG)~\cite{shaker2016procedural} means using algorithms to produce content. PCG techniques have long been utilized in video games, such as for level generation in Beneath the Apple Manor (Don Worth, 1978) and Rogue (Glenn Wichman, 1980). Today, PCG is widely used to generate content in games, such as maps in Spelunky (Derek Yu, 2008), terrain in Minecraft (Mojang Studios, 2009), and worlds in Starbound (Chucklefish, 2016) and No Man's Sky (Hello Games, 2016). Research in PCG is expansive; in this section, we review PCG techniques within the video game domain. Specifically, we describe research using search-based, machine learning, and reinforcement learning methods.

\subsection{Search-based Procedural Content Generation}
Search-based PCG defines a family of PCG techniques powered by search methods to find good content~\cite{togelius2011search}. In practice, evolutionary algorithms are often used, as they can be applied to many domains, including content and level generation in video games. Search-based PCG has been applied in many different game frameworks and games, such as the General Video Game AI framework~\cite{perez2016general},
Cut the Rope (ZeptoLab, 2010)~\cite{shaker2013evolving}, and generic mazes~\cite{ashlock2011search}. Search-based techniques have been used to generate levels~\cite{ashlock2011search,khalifa2015automatic,bhaumik2020tree}, board games~\cite{browne2010evolutionary} and level generators~\cite{kerssemakers2012procedural,khalifa2020multi}.

The process of training or evolving one or a population of level generators, as in the present work, can be seen as a form of SBPCG through the space of generators, where we search for a generator that is optimal in some content-generation task.

\subsection{Procedural Content Generation via Machine Learning}
Procedural content generation via machine learning~\cite{summerville2018procedural} (PCGML) is a family of PCG techniques that use machine learning algorithms to generate content. These methods are rarely seen outside of the research community because of their reliance on large datasets, long training times, and little control of the generated output. For example, Caves of Qud~\cite{grinblat2016caves} uses PCGML to generate books and other aesthetic elements. Research applications with PCGML have used many different techniques including Markov Chains~\cite{snodgrass2016learning}, N-Grams~\cite{dahlskog2014linear}, GANs~\cite{volz2018evolving,torrado2020bootstrapping}, Autoencoders~\cite{jain2016autoencoders}, and LSTMs~\cite{summerville2016learning}.

\begin{figure}
    \centering
    \includegraphics[trim={0 0 0 0},clip,width=0.49\linewidth]{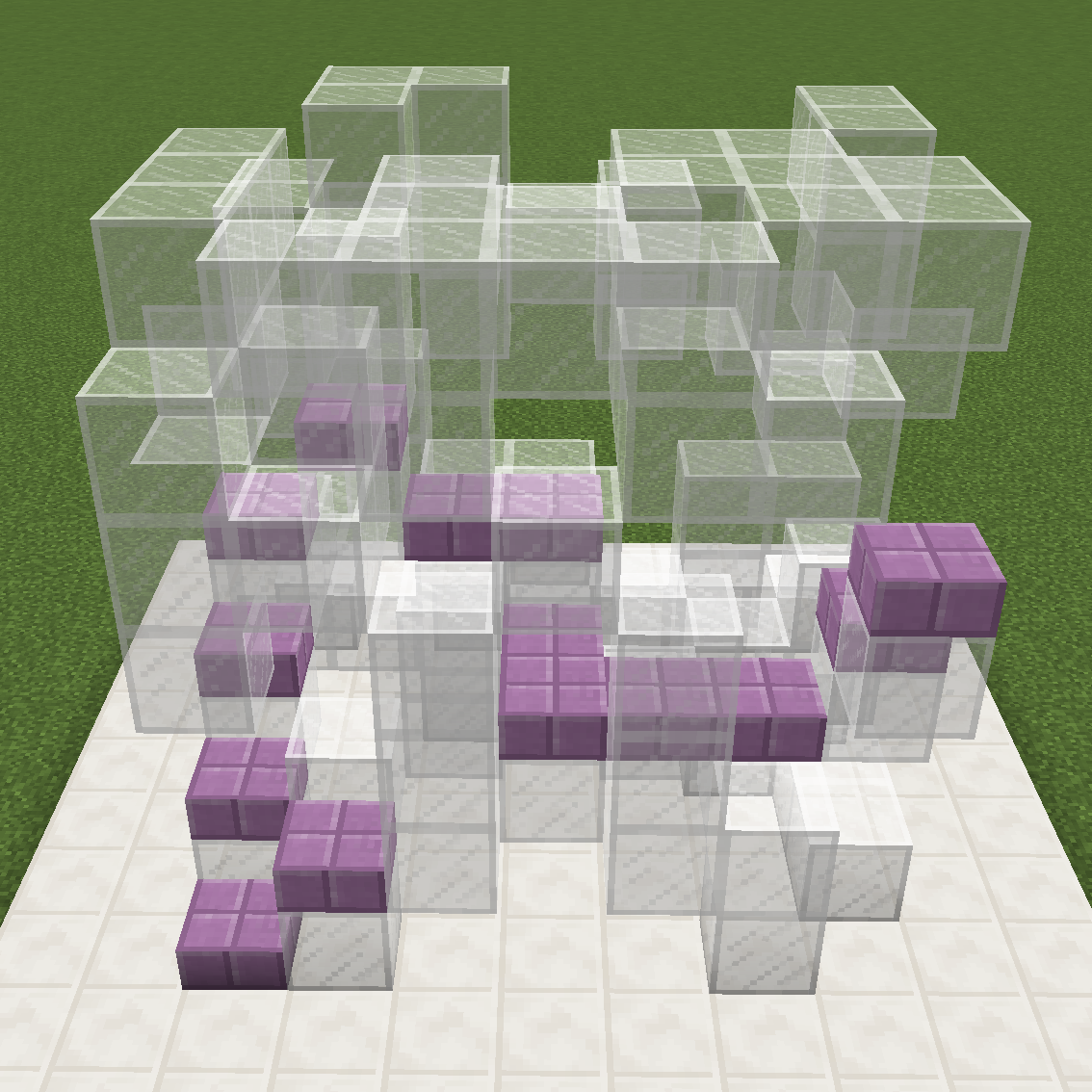}  
    \includegraphics[trim={0 0 0 0},clip,width=0.49\linewidth]{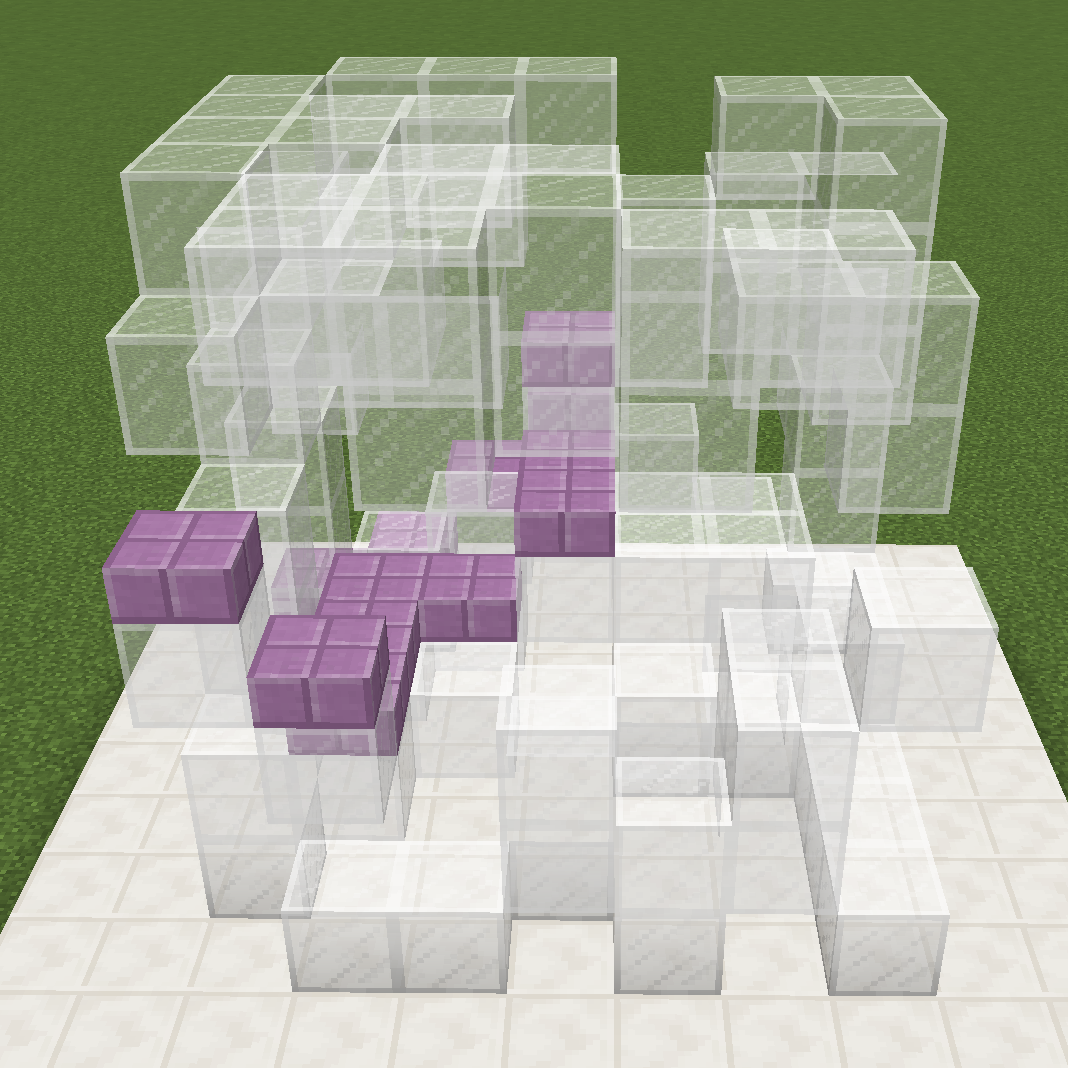}
    \includegraphics[trim={0 0 0 0},clip,width=0.49\linewidth]{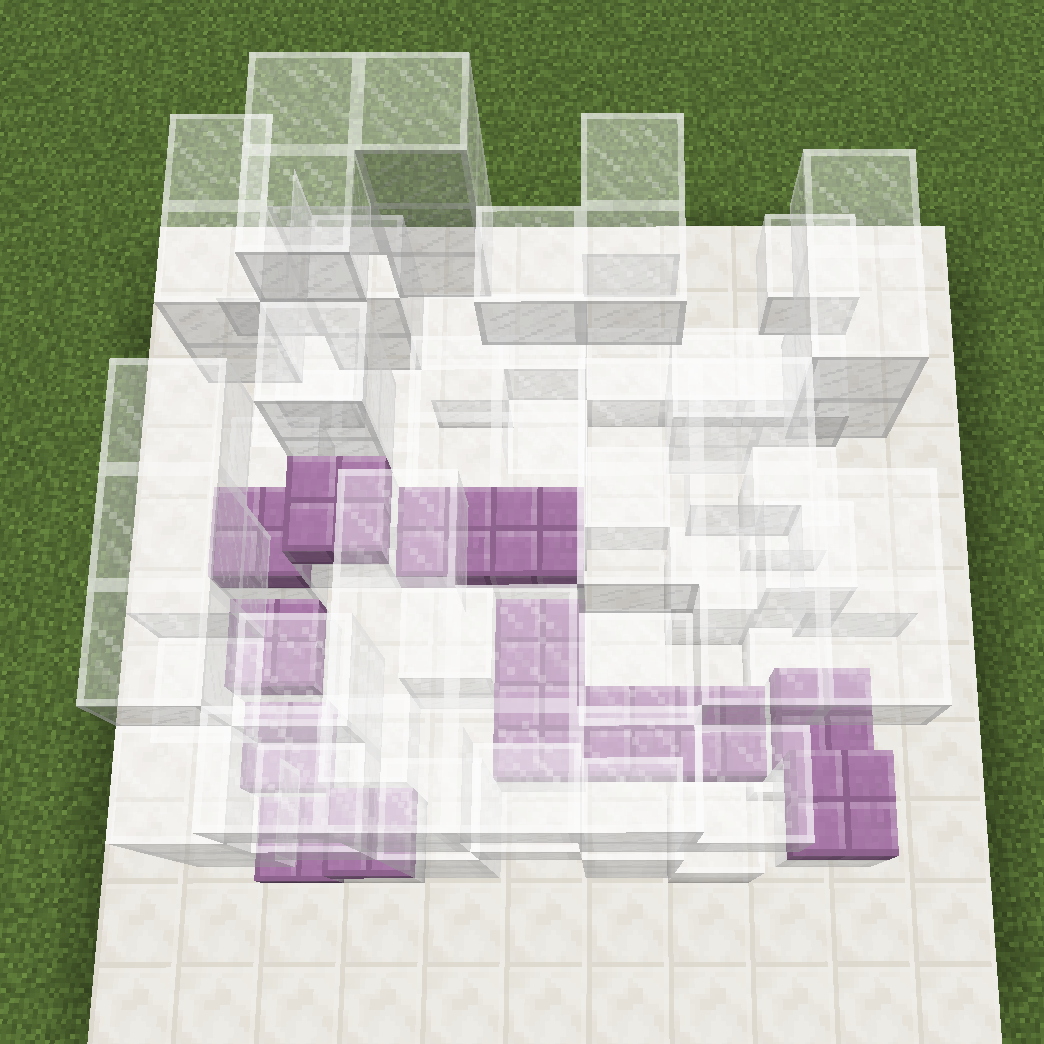}  
    \includegraphics[trim={0 0 0 0},clip,width=0.49\linewidth]{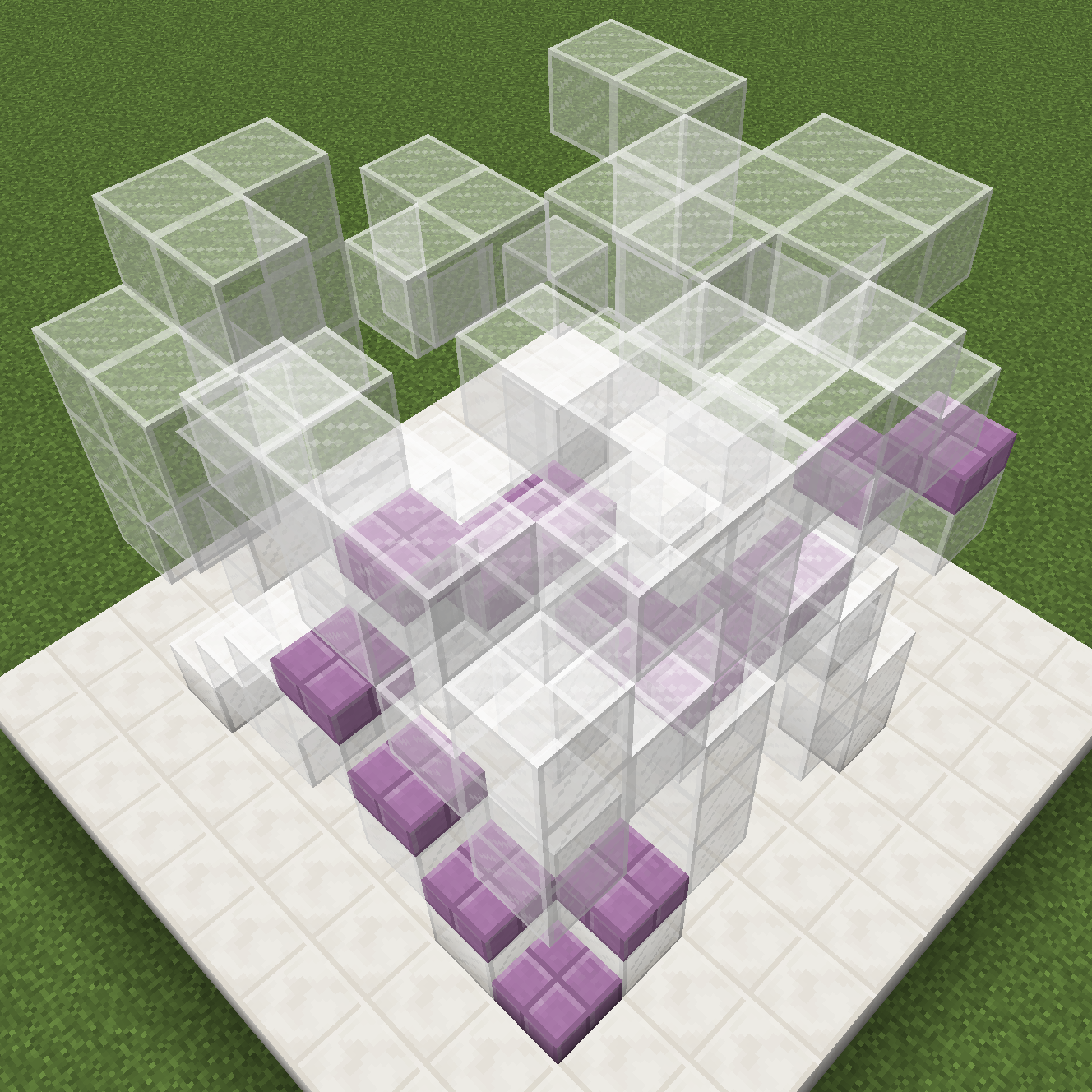}
    \caption{An example of the diameter task, in which an agent creates a maze in a $7 \times 7 \times 7$-voxel empty space, so as to maximize diameter (the longest shortest path between any two points) in the maze domain by constructing a winding staircase. Here we show the result of the agent's actions midway through a level-generation episode, from multiple angles.}
    \label{fig:diam_final}
\end{figure}

\subsection{Procedural Content Generation via Reinforcement Learning}
Procedural content generation via reinforcement learning~\cite{khalifa2020pcgrl} (PCGRL) is a family of PCG techniques powered by reinforcement learning algorithms. PCGRL techniques have primarily been used in level/experience generation~\cite{earle2021learning,
mahmoudi2021arachnophobia,chen2018q}. It is somewhat unintuitive to transform level generation into a reinforcement environment, as RL is more often used to play games. One way to do this in practice is to turn PCG into an iterative, co-creative process~\cite{guzdial2018co}.

In the original PCGRL formulation, diversity in a generator's output is ensured by initializing the level to a random state; after which the length of a generation-episode is constrained such that the generator may only modify a pre-specified percentage of the level. A follow-up work for training \textit{controllable} content generators allows the user to specify certain high-level metrics instead, which the generator should learn to control \cite{earle2021learning}.
The generator should then be able to produce levels approaching a certain value given as a target in any of these metrics.
This is achieved by sampling targets in each controllable metric throughout training (either randomly or using a curriculum based on agent learning progress), then feeding these targets to the agent as part of its observation and rewarding it for approaching these targets over the course of an episode.

The notion of user control is indispensable to developing generators with downstream co-creative applications.
In addition to high-level features, human users may also want to build alongside the agent, as in \cite{delarosa2021mixed}.
This motivates our extension of the notion of controllability to lower-level level features (individual tiles).
In addition to using the method in \cite{earle2021learning} to control aspects of the maze diameter task, we introduce two additional tasks in which user controls correspond to the placement of fixed structures in the level (i.e. doors), around which the generator must work to maximize level quality.

\subsection{Evolving Diverse Generators}
In addition to training generalist controllable agents, it is possible to evolve a large population of specialized generators, diversified along level metrics of interest. Such generators can, for example, be represented as cellular automata~\cite{earle2021illuminating}.
This method is similar to PCGRL in that it frames level generation as a Markov Decision Process, but uses Quality Diversity evolutionary algorithms instead of RL to search for generators.
It can also be seen as controllable, because a user could swap between models from the archive while designing levels.

Notably, these specialist generators are trained on relatively small batches of initial random levels, and are able to overwrite the entire initial state many times over.
We use this approach to search for diverse generators along high-level features of interest.
In the future, it could also be worthwhile to explore the ability of such models to generalize to user control over an unchangeable part of the initial state with a wide degree of possible variation (the entrance/exit doors of the level).






\begin{figure}
    \centering
    \includegraphics[trim={0 0 0 0},clip,width=0.49\linewidth]{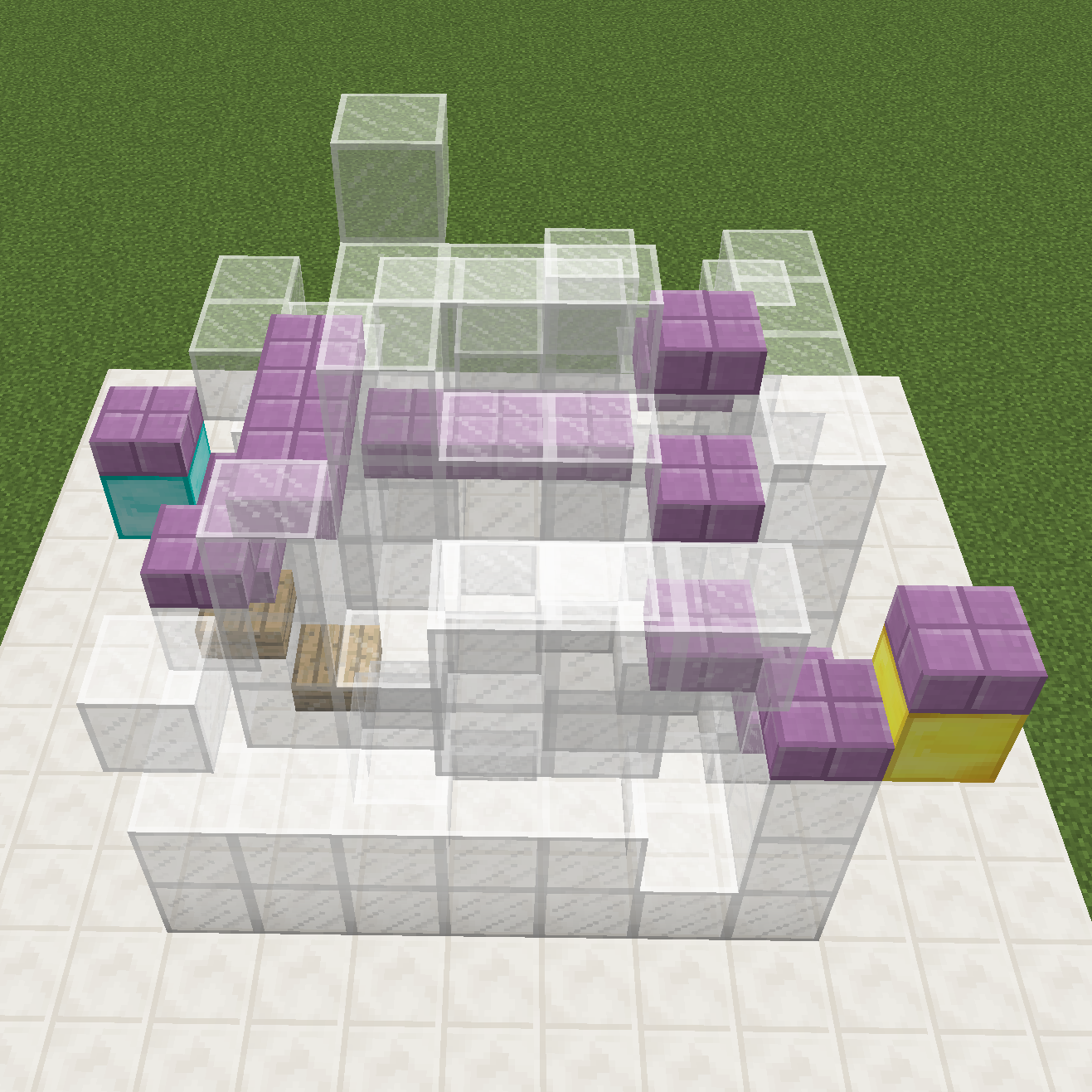}
    \includegraphics[trim={0 0 0 0},clip,width=0.49\linewidth]{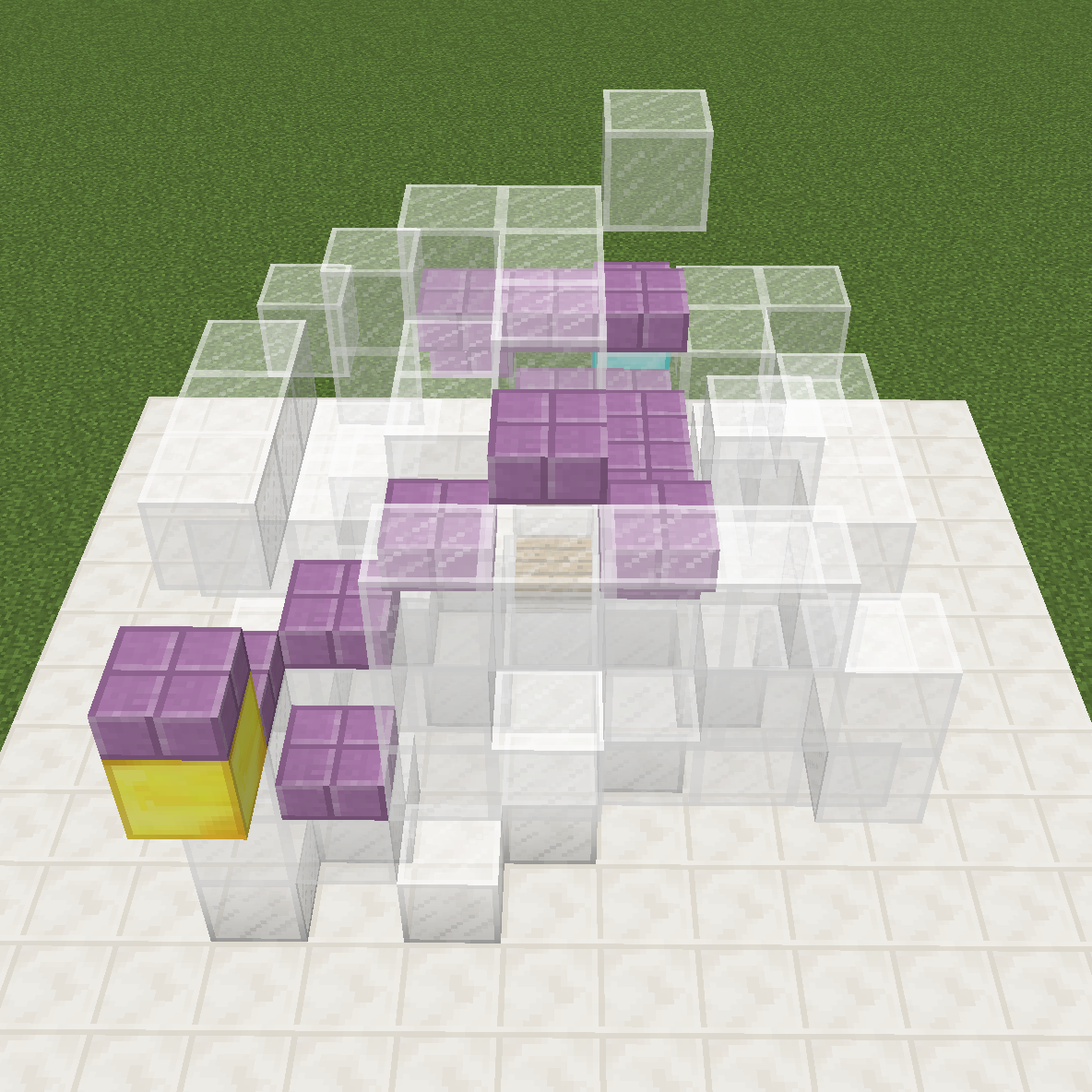} 
    \includegraphics[trim={0 0 0 0},clip,width=0.49\linewidth]{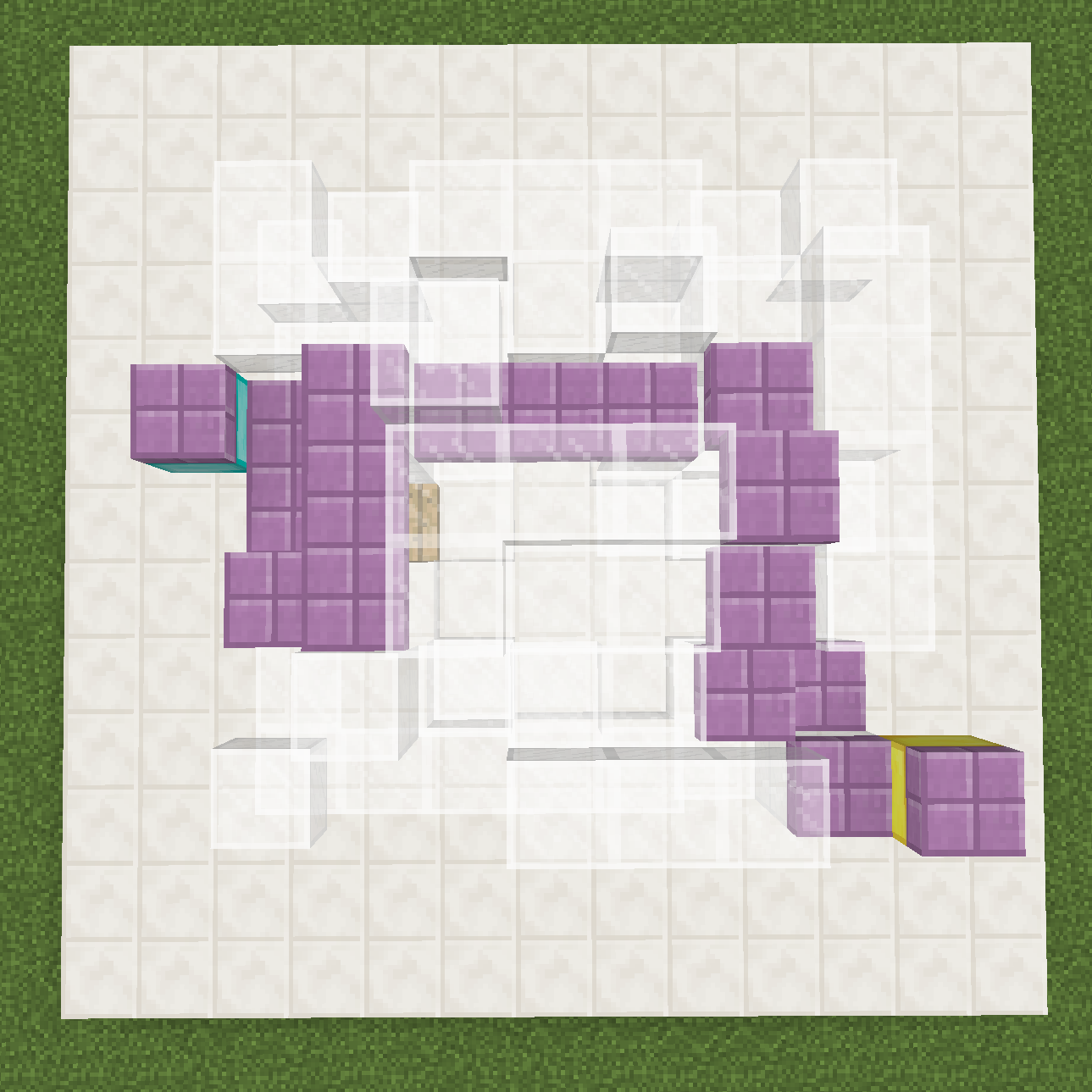} 
    \includegraphics[trim={0 0 0 0},clip,width=0.49\linewidth]{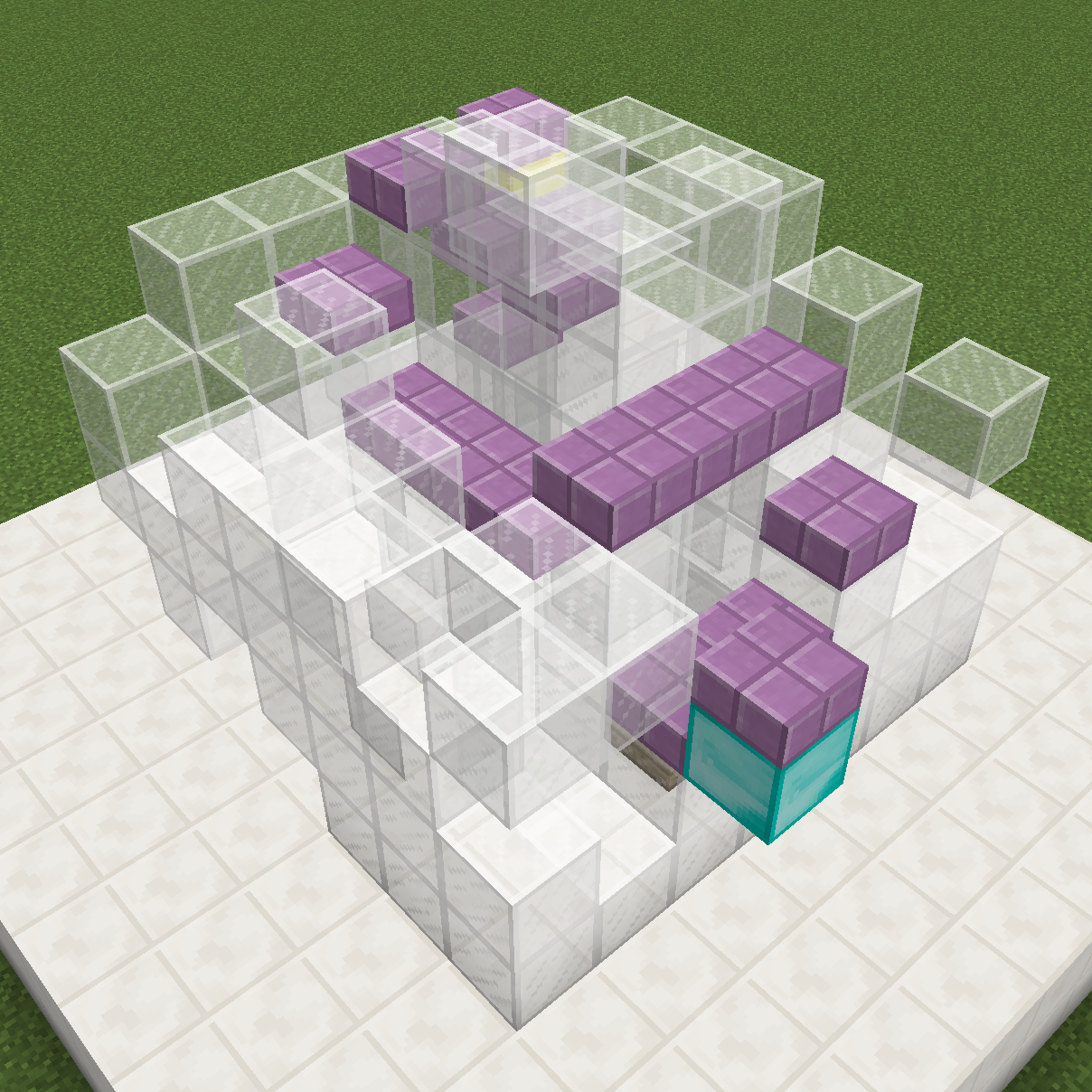} 
    \caption{An example of the connecting-doors task in which the agent connects the doors on the border with a maximally lengthy traversable path. The entrance/exit are represented by gold/diamond respectively. Since the doors are placed at approximately the same height, across from one another, the agent complexifies the path connecting them by adding ascending/descending stairs in addition to a detour along the $x, z$ plane.}
    \label{fig:door_final}
\end{figure}

\section{Methods}
In the following subsections, we explain the proposed 3D domain tasks and the generator model, adapted from the 2D PCGRL framework, and describe how we render experiment results using Evocraft \cite{grbic2021evocraft} for Minecraft.

\subsection{3D Tasks}
As in previous PCGRL studies~\cite{khalifa2020pcgrl}, we use different tasks to test the ability of our model in generating game levels. In this paper, we extend several tasks that have appeared in the 2D PCGRL domain to 3D. In all of them, the agent must build a maze within a $7\times 7\times 7$ cubic space which is initialized empty. The agent may observe the maze's borders and all the blocks within the space. At each step, the agent sequentially scans over a position inside the maze, and decides which type of block to place. Unlike image generation, a popular choice of task for ML generative systems, maze generation tasks have functionality constraints. Mazes must be traversable by embodied players usually while having an entrance and an exit. Also, 3D gravitational environments may have physical obstacles requiring jumps, where a jump is defined as a 
trajectory taking the player up, over some gap or obstacle, or down onto a solid platform which is not directly adjacent to the player's current position.

We introduce four maze generation tasks below:
\begin{enumerate}
    \item \textbf{Maximizing maze diameter}: the agent must create a maze with maximum diameter (the length of the longest-shortest path between any pair of player positions within the maze). The agent is rewarded for increasing the diameter and including a target number of jumps along this path.  
    \label{tsk:diam}
    
    \item \textbf{Connecting given doors}: the agent must connect given doors, fixed on the maze's borders at the start of each episode, with a maximally complex path. The agent observes the border (including the doors) but cannot change it. The agent is rewarded for maximizing path length and producing a target number of jumps occurring in the path. We define the path as the furthest path that can be reached from the entrance door. In order to encourage the agent to link the given two locations, when the agent connects the entry and exit points, we distinguish between the connected path and the furthest path that can be reached from the entry point, and give more weight to the connected path. We thus provide a denser reward during training, while ultimately encouraging the agent to focus on complexifying the path between doors.
    \item \textbf{Dungeon generation}: the agent must build a maze between a randomly placed entrance/exit as in the previous task, while also placing a specified number of enemies and a unique goal (chest) within the maze. The path of interest is that linking the entrance to the chest, plus that connecting the chest to the exit. The agent is rewarded for generating the right number of enemies, goals and jumps; maximizing the path length; and maintaining a minimum distance between the enemies and the entrance. 
    
    \item \textbf{Controlling high-level features}: given any of the above tasks, the generators are additionally asked to control high-level features of the level. 
    In this paper, we ask the generator to control the number of jumps and path-length of the diameter in task~\ref{tsk:diam}.
    
\end{enumerate}



During training in the door connection and dungeon generation tasks, the agent is given random entrance/exit locations at the beginning of each episode. However, these are fixed during evaluation to create comparable results. 

\subsubsection{3D pathfinding}

We consider the cube above that on which the supposed player agent is standing as the agent's current position (its foot-room).
The cube above that is the player's head-room, and it must be empty.
In computing our path-related metrics for generated levels, we allow the following actions to comprise a path.
The player can move to adjacent cubes along $x$ or $z$ at the same height ($y$) if the foot- and head-room tiles are passable and the tile below the foot is solid.
The player can move up one ``stair'' provided an extra cube of head-room above the player, a solid cube as new support beside the current foot-room, and head- and foot-room above the new support. Or, the player can move down one ``stair'' provided there are passable tiles beside its current head-room, foot-room, current foot support, and solid support below the next potential foot-room. 
Note that the extra room will be required when the player goes either up or down a tile along the $y$ axis.
The player can ``jump'' to a cube two tiles along $x$ or $z$, and either up or down 1 in $y$, provided a ``gap'' beside the player. A gap is present when at least 2 tiles below an adjacent position are empty. 
Similar to stair-climbing, there must be passable cubes for headroom at the $y$ position above the player's head, both at the player's current $(x, z)$ position, and at that of the gap.
The two cubes above the $(x, z)$ position where the player will end up after jumping must also be passable to accommodate the player.
For the sake of simplicity, we do not include any falls/jumps involving more drastic changes in the player's height, in order to ensure that paths are re-traversable.

\subsubsection{Reward specifications}

Following the original PCGRL framework, at each time-step in an episode of level-generation we compute the distance between the current and target value in each metric of interest of the current maze. In the diameter-maximization and door-connection tasks, this means we compute the number of jumps and path-length of the relevant paths. In the dungeon task we additionally compute the number of chests, enemies, and the shortest path from the entrance door to the nearest enemy.
We call this the ``loss'' $l_t$ of the currently-generated level.
At the next time-step, we compute the update's loss $l_{t+1}$ of the maze.
The agent's reward is then given by $R_{t+1} = l_{t+1} - l_t$.

For each task, we define a set of control metrics $M \def \left{m_0, m_1, \dots, m_i,\dots m_n\right}$.
\begin{table}[]
\begin{tabular}{l|l|l|l}
\textbf{Task}    & \textbf{Metrics $(m_i)$} & \textbf{Weights $(w_i$)} & \textbf{Targets $(g_i)$}     \\
\toprule
Diameter         & n. jumps         & 1       & 5           \\
                 & diameter         & 1       & $\infty$      \\
\hline
Connecting doors & n. jumps         & 1.5     & 5           \\
                 & diameter         & 1       & $\infty$      \\
                 & path length      & 1.2     & $\infty$      \\
\hline
Dungeon          & n. jumps         & 1       & $[2, 5]$  \\
                 & n. chests        & 3       & 1           \\
                 & n. enemies       & 1       & $[2, 5]$  \\
                 & nearest enemy    & 2       & $[5, \infty)$ \\
                 & path length      & 1       & $\infty$     
\end{tabular}
\caption{Metrics, weights, and targets for each task. In the connecting-doors problem, path length measures the shortest path between doors and diameter measures the longest shortest path from the entrance. In the dungeon problem, path length measures the shortest path from entrance, to chest, to exit door.}
\end{table}
At each timestep $t$ in the level-generation episode, each metric $m_i$ takes on value $m_{i, t}$ depending on the state of the level.
(For example, in the diameter problem, an empty $7$-voxel-wide cube would result in a path length of $14$.)
At the beginning of each episode, we also fix a target for each metric, $g_{i}$.
Then at each timestep we define the level's design loss as
\begin{align*}
l_t &= \sum_i w_i \lvert \hat{g}_i - m_{i, t}\rvert
\end{align*}
, where $w_i$ is an importance weight on metric $i$.
When the target $g_i = \infty$, we set $\hat{g}_i = \min(m_i)$, the minimum possible value of $m_i$, so as to encourage maximizing the metric.
When the target $g_i$ is an interval, we let $\lvert \hat{g}_i - m_i\rvert$ be the minimum distance between $m_i$ and any value in the interval.


We choose the above ``dense'' reward scheme, in which loss is computed at each time-step, as opposed to a sparse reward scheme in which it is only computed at the end of an episode, because it is anecdotally found to result in lower sample complexity (i.e. agents require fewer frames of training before learning interesting behaviors), though it may be liable to result in more ``greedy'' learned behaviors.

We choose to define reward as the difference in losses between time-steps, as opposed to the loss itself, because the latter scenario would never result in positive reward, which could result in more conservative learned behavior from the agent. 
Having motivated these design choices, we note that both could merit further exploration and empirical analysis in future work.

\begin{figure}
    \centering
    \begin{subfigure}{1\linewidth}
    \includegraphics[trim={0 0 0 20},clip,draft=false,width=1.0\linewidth]{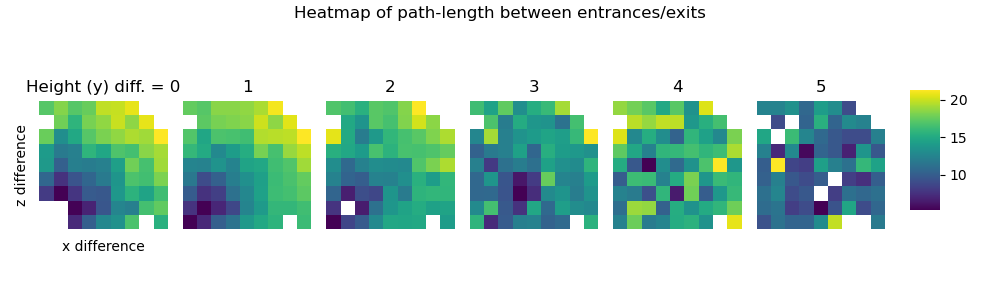}
    \caption{Mean path-length between connected doors.}
    \end{subfigure}
    \begin{subfigure}{1\linewidth}
    \includegraphics[trim={0 0 0 20},clip,draft=false,width=1.0\linewidth]{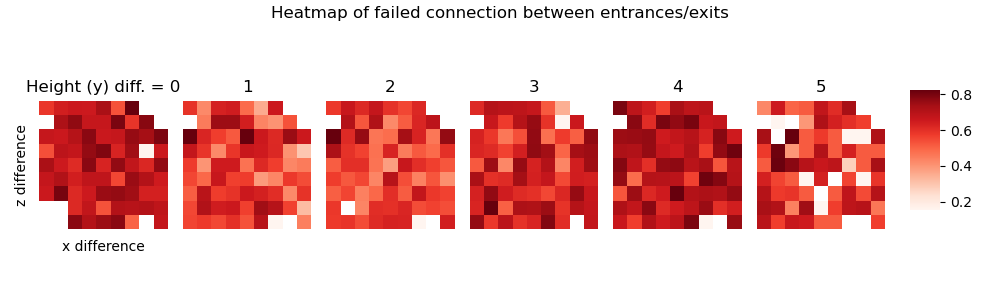}
    \caption{Rate of failure (disconnected paths) between doors.}
    \end{subfigure}
\caption{Agent performance in connecting all possible pairs of random doors. We average over symmetric relationships between doors, by considering the absolute distance between the doors in each dimension}
\label{fig:hole_heat_distance}
\end{figure}

\begin{figure}
    \begin{subfigure}{.49\linewidth}
    \includegraphics[trim={40 0 40 30},clip,draft=false,width=1.0\textwidth]{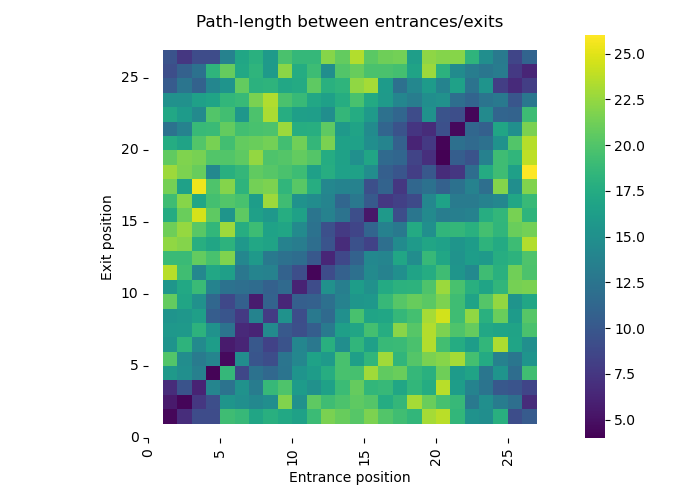}
    \caption{Mean path-length between connected doors.}
    \label{fig:hole_heat_unravel_scc}
    \end{subfigure}
    \begin{subfigure}{.49\linewidth}
    \includegraphics[trim={40 0 40 30},clip,draft=false,width=1.0\textwidth]{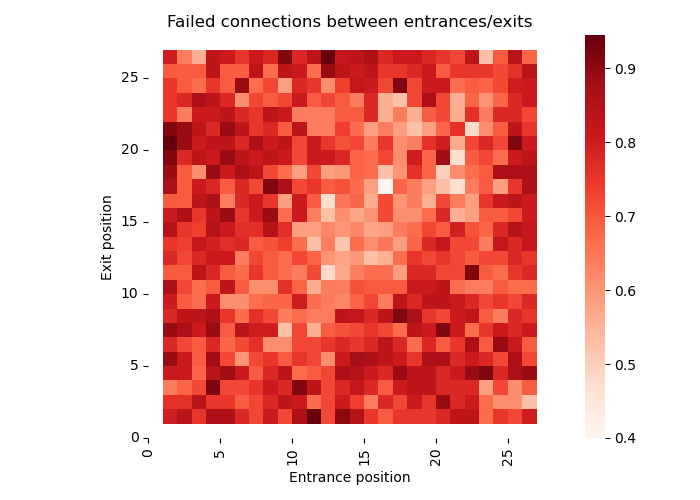}
    \caption{Rate of failure (disconnected paths) between doors.}
    \label{fig:hole_heat_unravel_fail}
    \end{subfigure}
    \caption{Agent performance in connecting all possible pairs of random doors. We consider the position of each doorway along the circumference of the maze, averaging over results at varying heights. Left: the mean path-length achieved when pairs of doors are connected. Right: the number of failures in connecting pairs of doors.}
    \label{fig:hole_heat_unravel}
\end{figure}

\subsection{Reinforcement learning generators}
We use Proximal Policy Optimization (PPO~\cite{schulman2017proximal}) to train our agent, with an architecture adapted for 3D from the original PCGRL framework. This consists of a 3D convolutional layer, followed by dense layers, with a separate branch each for action and value prediction. 
The 3D convolutional layer has $64$ hidden channels, a $3\times 3$ kernel, a stride of $1$ and padding of $1$.
The resulting activation is flattened 
and passed through a fully connected layer into an activation of size $64$.
The network then splits into separate action and value branches, with the former consisting of a fully connected layer to $n_a$ outputs, where $n_a$ is the size of the agent's action space (e.g. 2 in the diameter and door-connecting tasks, corresponding to placement of air or solid cubes), and the latter a fully connected layer to $1$ output (the value estimate).
The resulting models have $\approx22.5$, $27.7$, and $27.7$ million learnable parameters on the diameter, door and dungeon tasks, respectively.

We one-hot encode the 3D level as a 4D array---giving each tile type its own channel---and padding the map so that the agent can fully observe the level while staying at the ``center'' of the observation. 
We also encode the current path as a separate one-hot encoded channel in the agent's observation. (In the diameter task, this path is that corresponding to the diameter of the current maze, while in the door-connection and dungeon tasks, this path is the shortest path connecting doors and/or chest, respectively, if such a path exists.) 
Unlike previous 2D studies in which the same material is used for padding as is used to represent the border, we distinguish padded from border tiles by filling the padded region with all-zeros.

\begin{figure}
    \centering
    \includegraphics[trim={0 0 0 0},draft=false,clip,width=0.49\linewidth]{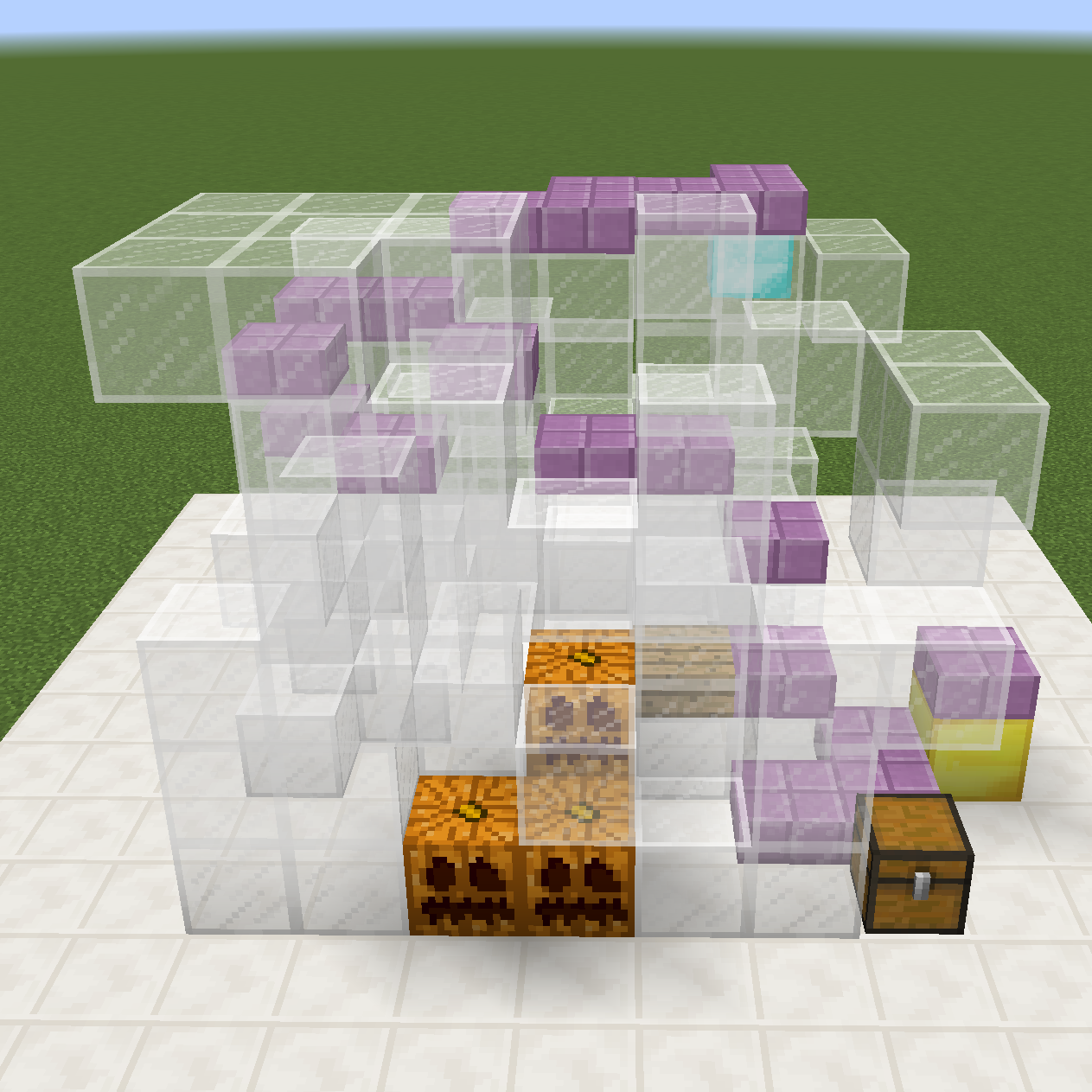}
    \includegraphics[trim={0 0 0 0},draft=false,clip,width=0.49\linewidth]{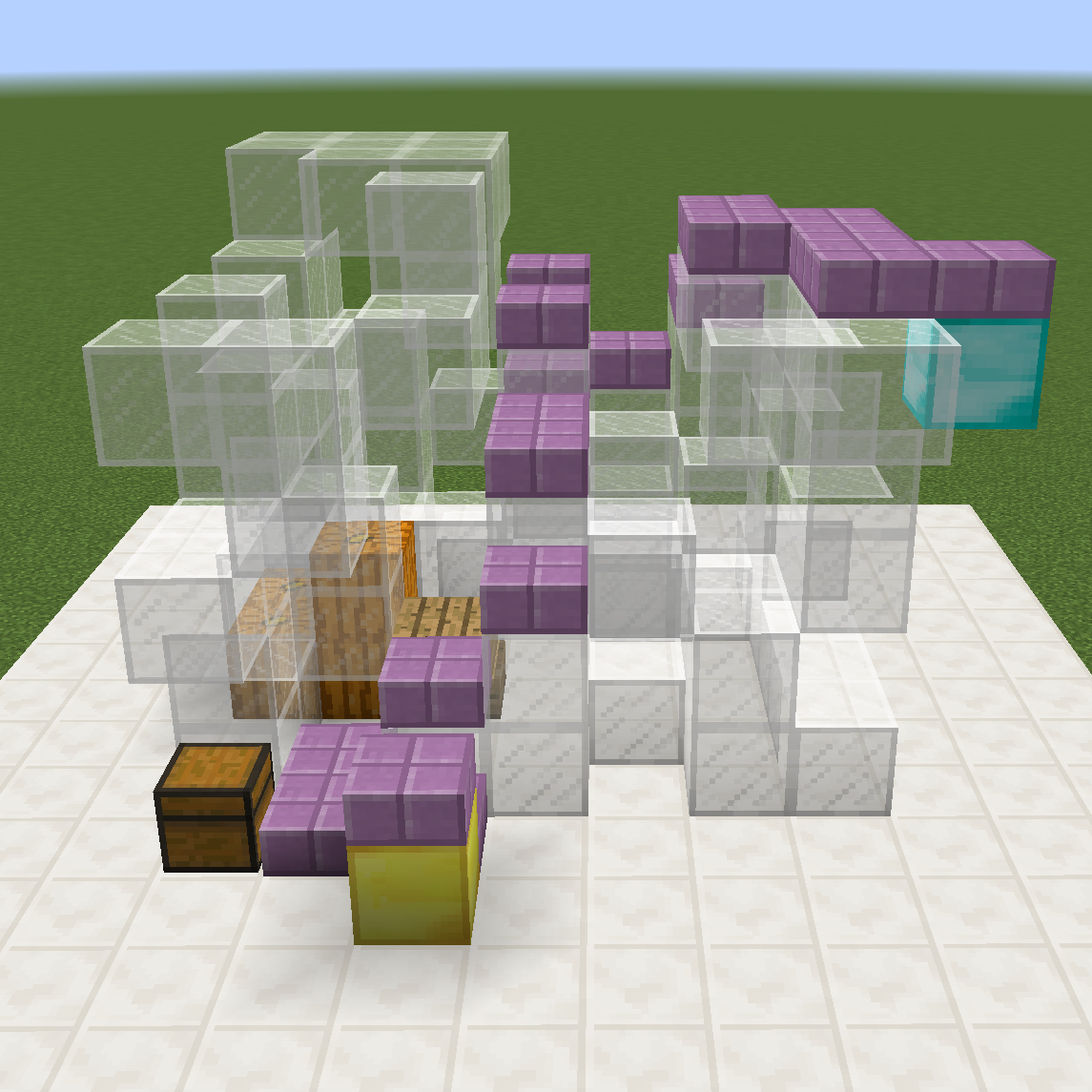}
    \includegraphics[trim={0 0 0 0},draft=false,clip,width=0.49\linewidth]{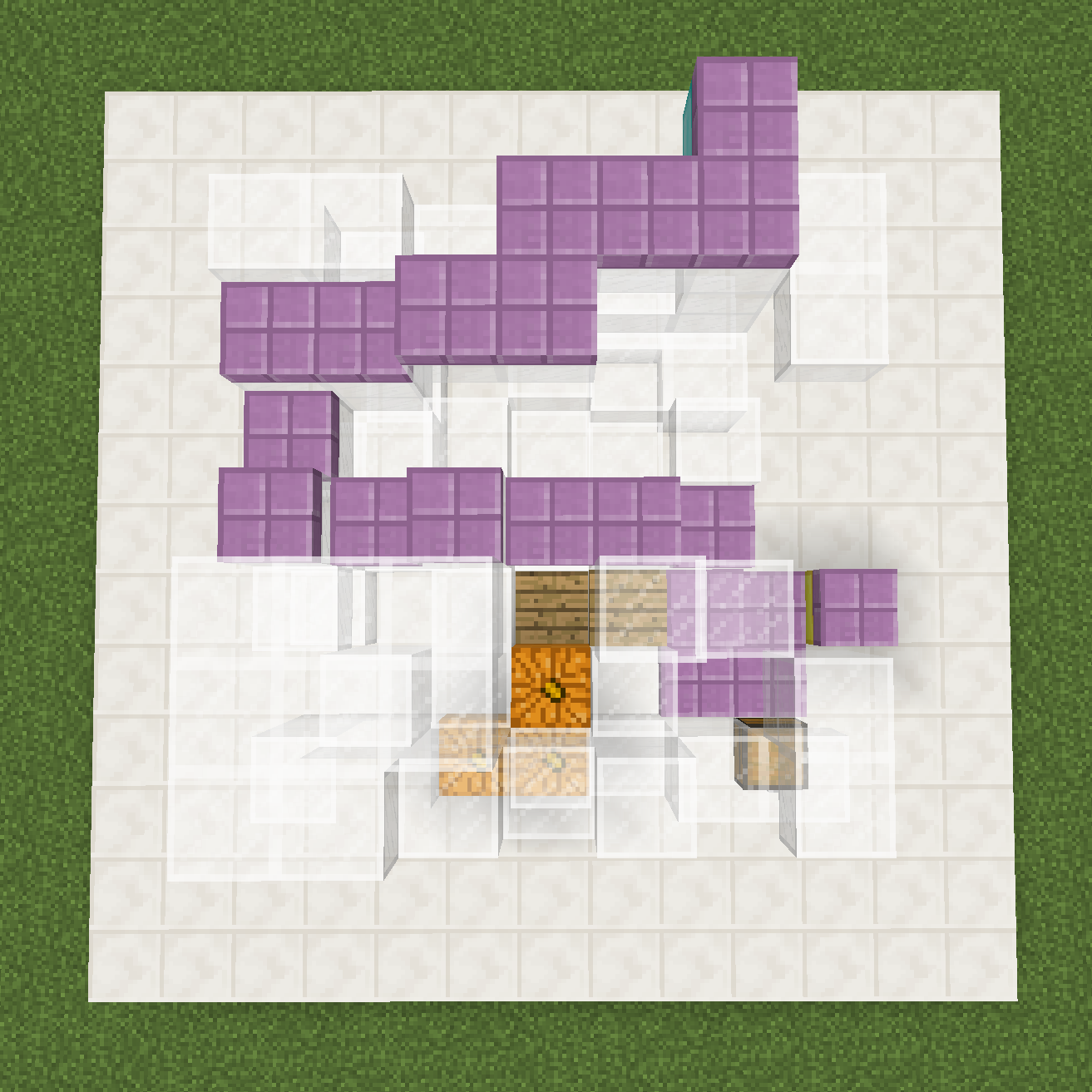}  
    \includegraphics[trim={0 0 0 0},clip,width=0.49\linewidth]{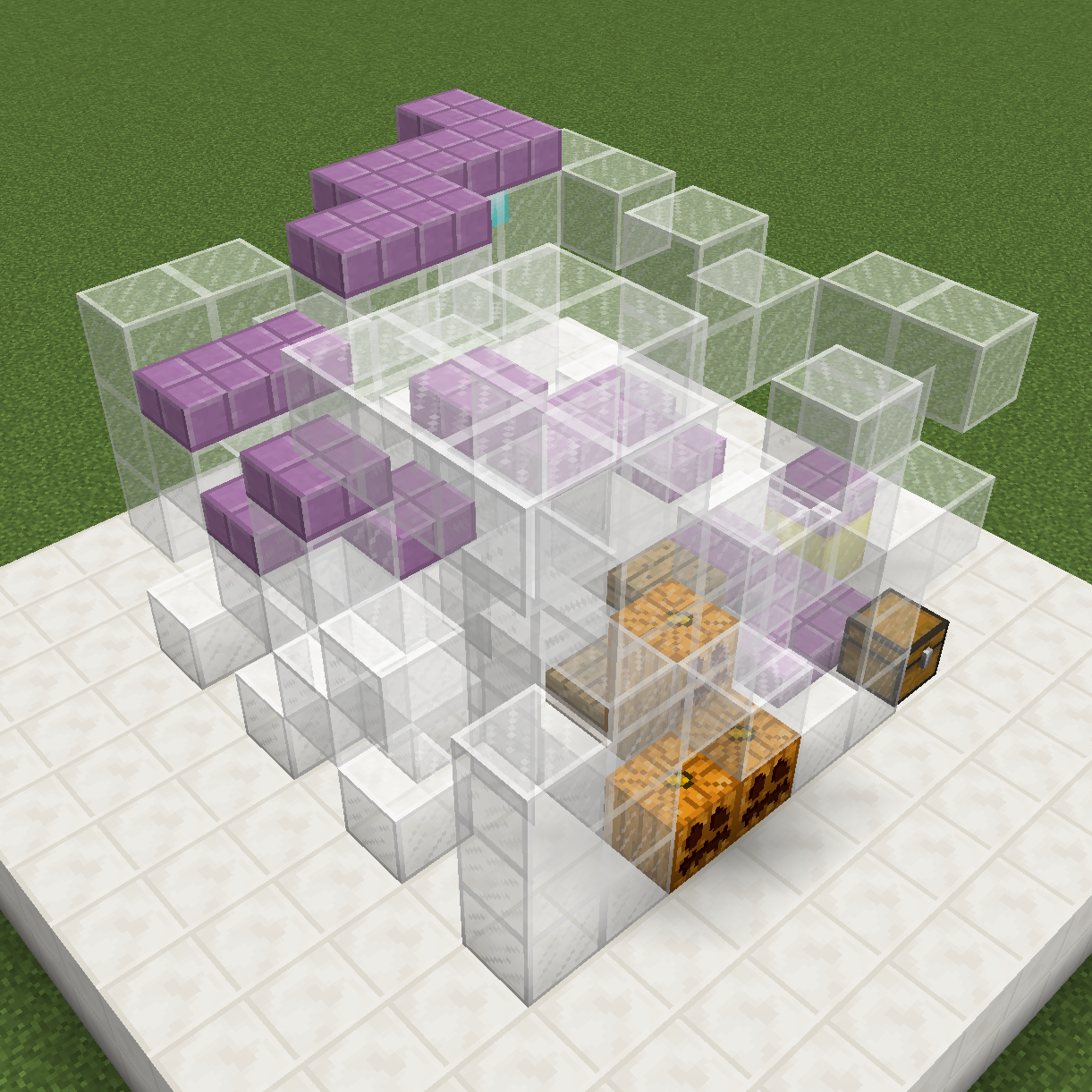}
    \caption{A generator produces a dungeon with a roundabout path from entrance, to chest, to exit. It places a moderate number of enemies (pumpkins) while keeping them at a minimum distance from the dungeon's entrance.}
    \label{fig:dung_final}
\end{figure}

\subsection{Evolving diverse generators}

To evolve diverse generators, we follow the method in \cite{earle2021illuminating}, adapting their Neural Cellular Automata (NCA) architectures to use 3D convolutions.
In this method, at each step, the generator observes the entire map and (potentially) edits each tile in response.

The model consists of three convolutional layers. The first convolutional layer has a $3 \times 3$ kernel, a stride of 1 and padding of 1, followed by a ReLU activation layer. The second layer has a $1 \times 1$ kernel, a stride of 1 and padding of 0, also followed by a ReLU activation layer. The last layer has a $1 \times 1$ kernel, a stride of 1 and padding of 0, followed by a sigmoid activation layer. The number of hidden channels is 32, and the input and output size of the model is the same as the map size.
These networks have $\approx2$, $6$, and $8$ thousand parameters on the diameter, door, and dungeon tasks, respectively.
Covariance Matrix Adaptation MAP-Elites \cite{fontaine2020covariance} is used to evolve these NCAs, resulting in generators that are diverse along measures of emptiness (the proportion of air cubes in the maze) and diameter.
Each NCA is evaluated on $10$ initial random level layouts. The initial layout is fed into the network, which then iterates on the level for $50$ steps, potentially changing the entire level at each step.
The fitness of the generator is calculated using the state of the $10$ levels after $50$ steps, using (in order of importance) the validity of the levels (here their adherence to the target number of jumps), their reliability (the inverse of their variance along measures of emptiness and path-length), and diversity (in terms of hamming distance) among themselves.

\subsection{Rendering in \textit{Minecraft}}

We use Evocraft to render our environment~\cite{grbic2021evocraft}. 
We use the ``red glazed terracotta'' block to render path-edit animations for clarity.  
The ``trapdoor'' block is used to render the path (due to its being traversible for human players when testing the generated levels).
The ``chest'' block is used to signify a key objective for players within a dungeon, while the ``pumpkin'' block is used to signify enemies (which could be replaced with creep-spawning locations at the start of play). In order to see the internal structure of the 3D game level more clearly, we use the ``white stained glass'' block to render solid tiles in the maze. 

\section{Results}
In the following subsections, we present results on the four 3D level-generation tasks: maximizing maze diameter, connecting fixed doors, generating dungeons, and controlling high-level maze features.

\subsection{Maze diameter}
To maximize the diameter of a maze, the agent first increases the initial path (spanning from one corner to another on the ground floor) by adding 
obstacles to the ground floor of the maze,
and forcing the shortest path between corners to climb and/or descend ``steps'' along the way.
(The most successful of the $3$ agents trained on this task produces a distinctly checkerboard-like pattern, which forces the shortest path to resemble a sawtooth, traversing redundant, adjacent stairs.)
The agent then extends this pattern of jagged cubes upward, creating a mountain-like shape, so that the longest shortest path involves spiraling upward around this mountain.

Because we initialize the maze to be entirely empty (the level is filled with ``air'' cubes), we can expect the agent to generate the same maze with the same sequence of actions when taking the maximum-likelihood action at each step.
Further, because there is presumably only one (or a handful) of optimal, maximum-diameter mazes inside a cube given our path-finding logic, we would expect a ``perfect'' generator to converge to producing the same level each episode.
Our trained RL agent on the maximum-diameter maze task indeed produces a similar pattern each episode, though this layout is far from optimal (i.e. it is easy to see how we could extend the path in Fig.~\ref{fig:diam_final}).


\subsection{Random door placement}

When the task has randomly-placed entrances and exits (along the 4 bordering walls of the maze), the agent similarly builds a central, staircase-like structure (Fig.~\ref{fig:door_final}).
This climbable structure requires only minor modifications to yield a traversable path between pairs of entrances/exits along the bordering walls, and subsequently, to render this traversable path more lengthy.

In Fig.~\ref{fig:hole_heat_unravel}, we iterate over all valid possible pairs of entrance/exit positions, evaluating the path-length the model manages to generate (between doors) as a result. 
Exploration is turned off (i.e. the maximum-likelihood action output by the policy is always chosen), and each pair of doors is evaluated for one episode, starting from an empty level as during training.
For the sake of visualization, we collapse the height dimension, and consider each door as being placed along a line covering the circumference of the maze.

From the dark line at $y=x$ in Fig.~\ref{fig:hole_heat_unravel_scc}, indicating lower mean path-length, we can see that the agent has consistent difficulty in constructing lengthy paths when attempting to connect doorways directly on top of one another.
At the same time, the bright spot around $x\in[15, 20], y\in[15, 20]$ in  Fig.~\ref{fig:hole_heat_unravel_scc} indicates that the generator is least likely to fail in connecting pairs of doors in this particular section of wall (even when these doors are stacked height-wise).
We conclude that nearby or vertically-stacked doors are relatively easy to connect (e.g. with a compact staircase), though the generator is more likely to connect these doors with a simple rather than complex path.

The bright bands on either side of $y=x$, on the other hand, suggest that the agent tends to produce longer paths when doors are diagonally across from one another in the width and depth dimensions.
In such cases, when both doors are close to the ground, the initial empty maze has a relatively long path-length, which the generator is then able to further extend by placing obstacles as in Fig.~\ref{fig:door_episode}.
Overall, the agent is able to connect a wide variety of door-placements, sometimes with complex paths in spite of the explosive number of possible such placements and the difficulty in accounting for vertical differences between the doors.

Fig.~\ref{fig:door_final} would seem to provide a counter-example to this phenomenon, in which the agent connects vertically-stacked doors using a lengthy path.
But this relatively high-reward level layout occurs in the middle of the training episode, being broken at a later point during the episode.
We conclude that while is is possible to generate complex connecting paths for nearby or vertically-stacked doors, such configurations are fragile.

Similarly, in Fig.~\ref{fig:hole_heat_distance}, we iterate over all door pairs and record the length of the resulting paths.
Here, instead of averaging over the height ($y$) dimension, we collapse pairs of doors which are symmetric (i.e. equal given some reflection(s) in any dimension).

We note that empty (white) cells in the heatmaps correspond to door-pairings that would not be possible, or would not allow for non-trivial paths to be constructed, and briefly describe every such case here for completeness. Assume for convenience that $\max{x} = \max{y}$ and let $\max{x}\coloneqq w$ (here, $w=6$). When all of $x,y,z\in[0,1]$, the doors are either overlapping or share edges in such a way as to prohibit non-trivial paths. When $x = z = w$, the doors would have to be in opposite corners, with neither sharing a face with the interior of the maze, which is invalid. When $x=w$ and $z=w-1$ or vice versa, one of the two doors would have to be in such a corner. When $x=w-1$ and $z=0$ or vice versa, either one door would have to be inside the maze, or the opposite door would again have to be at an invalid corner. 

In Fig.~\ref{fig:hole_heat_distance}, again, we note that generated path-lengths tend to be less when doors are very close to one another (i.e. in the lower-left corners on heatmaps with $y\in [1, 2]$).
Conversely, when doors are far apart from one another in both the $x$ and $z$ dimensions, with little height difference, paths tend to be long (interestingly, this does not just apply to doors on the ground floor, but also to doors that are both high above the ground, in which case the generator can often be observed to start by building a simple elevated ``floor'' at the required height).
We note that as the height difference between doors increases, the generator's performance becomes far less reliable and consistent given changing $x$ and $z$ differences.
This suggests that the task of building stairs in order to account for height differences poses a unique challenge to the model, and coould lay the groundwork for fruitful future research.

\subsection{Dungeon generation}

In constructing a dungeon, the agent learns to build a single chest, normally in a corner of the maze, while limiting the number of placed enemies (Fig.~\ref{fig:dung_final}).
Again, a climbable structure is constructed such that the chest is serviced by multiple pathways to potential entrances and exits.
Often, a one-way path to the chest from entrance/exit is constructed before the opposite exit/entrance is connected.
The agent is prone to re-use large sections of path between chest and entrance/exit, as this maximizes total path-length while requiring no extra space.

\subsection{Controllable level metrics}

\begin{figure}
    \centering
    \includegraphics[trim={0 0 0 40},clip,draft=false,width=.8\linewidth]{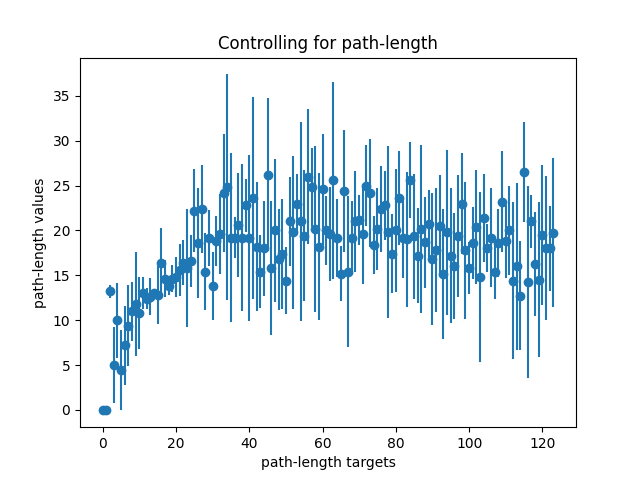}
    \caption{A generator trained to control the length of the diameter of a maze. The generator has good control over lower path-lengths, but is unable to attain very long paths when prompted.}
    \label{fig:ctrl_path}
\end{figure}

In Fig.~\ref{fig:ctrl_path}, we show the behavior of an agent trained to produce mazes having a diameter with a target length.
(While attempting to produce target path-lengths, the generator also aims to produce a static target of $5$ jumps.)
We see that for lesser path-lengths (between $0$ and $25$), the generator responds clearly to the user's input, generating an increasingly lengthy path as the user's target input increases.
For higher targets, however, the mean path-length of levels produced by the generator plateaus, and the variance of these path lengths increases, indicating that it has not learned to reliably control for mazes with large diameter.

This discrepancy could possibly be remedied by further training, or more efficient forms of training, for example using a curriculum for sampling control targets on which to train, a specialized network architecture (e.g. with separate value heads for different components of the reward function), or other algorithmic modifications.
Training to control for the number of jumps did not result in a significantly controllable agent, and we thus leave this as an additional challenge for future work.

\begin{figure}
    \begin{subfigure}{1\linewidth}
    \centering
    \includegraphics[trim={0 0 0 0},clip,width=.49\linewidth]{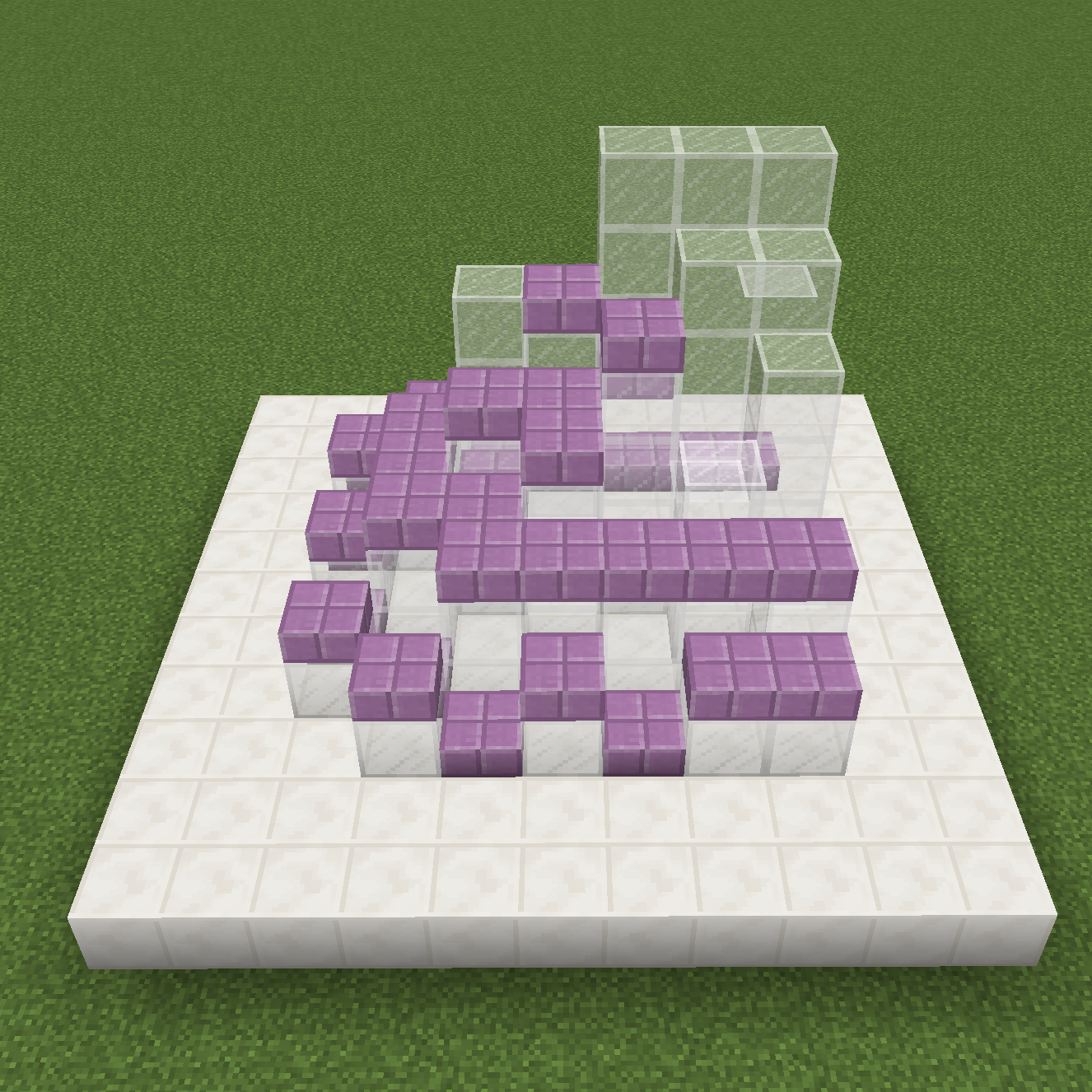}
    \includegraphics[trim={0 0 0 0},clip,width=.49\linewidth]{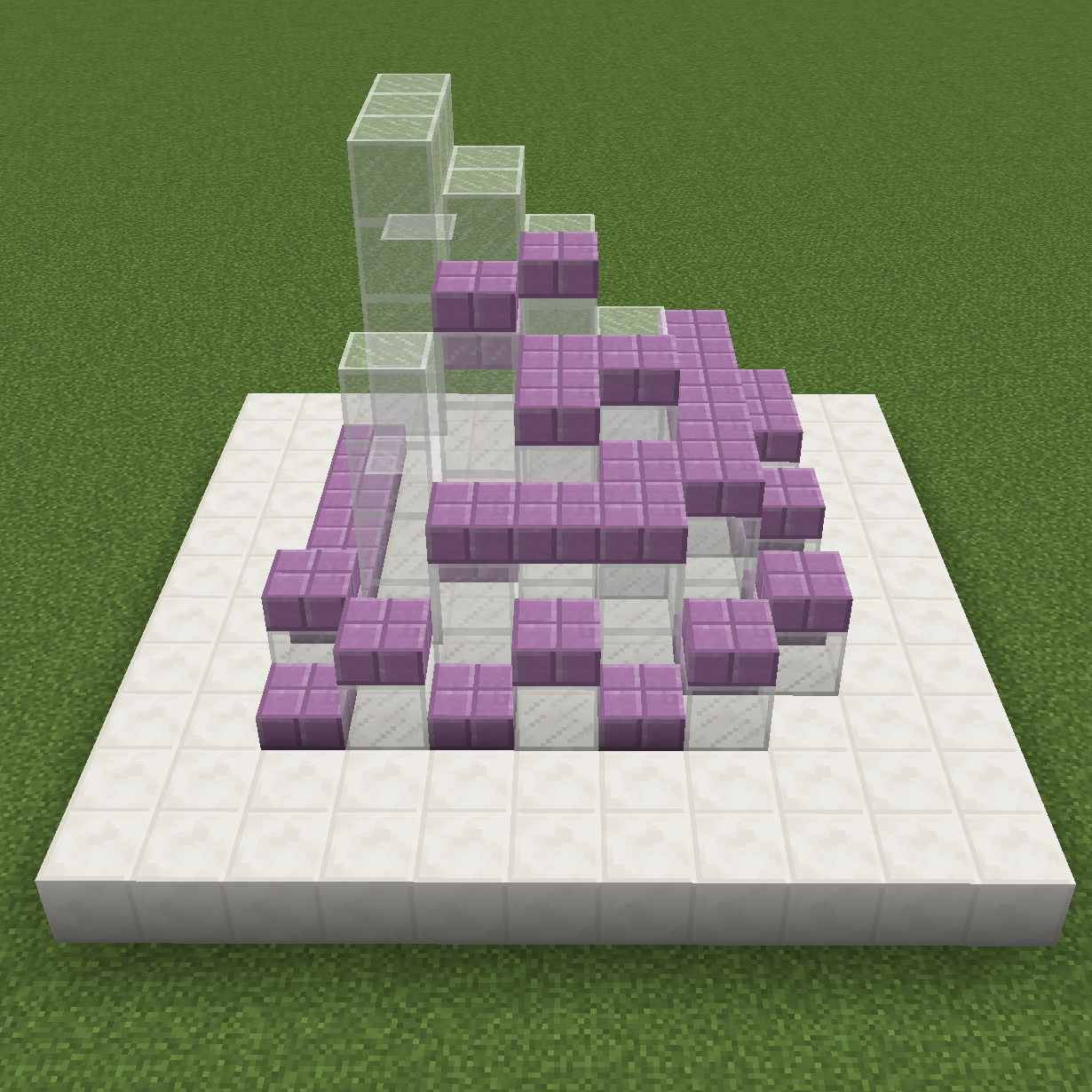}
    \includegraphics[trim={0 0 0 0},clip,width=.49\linewidth]{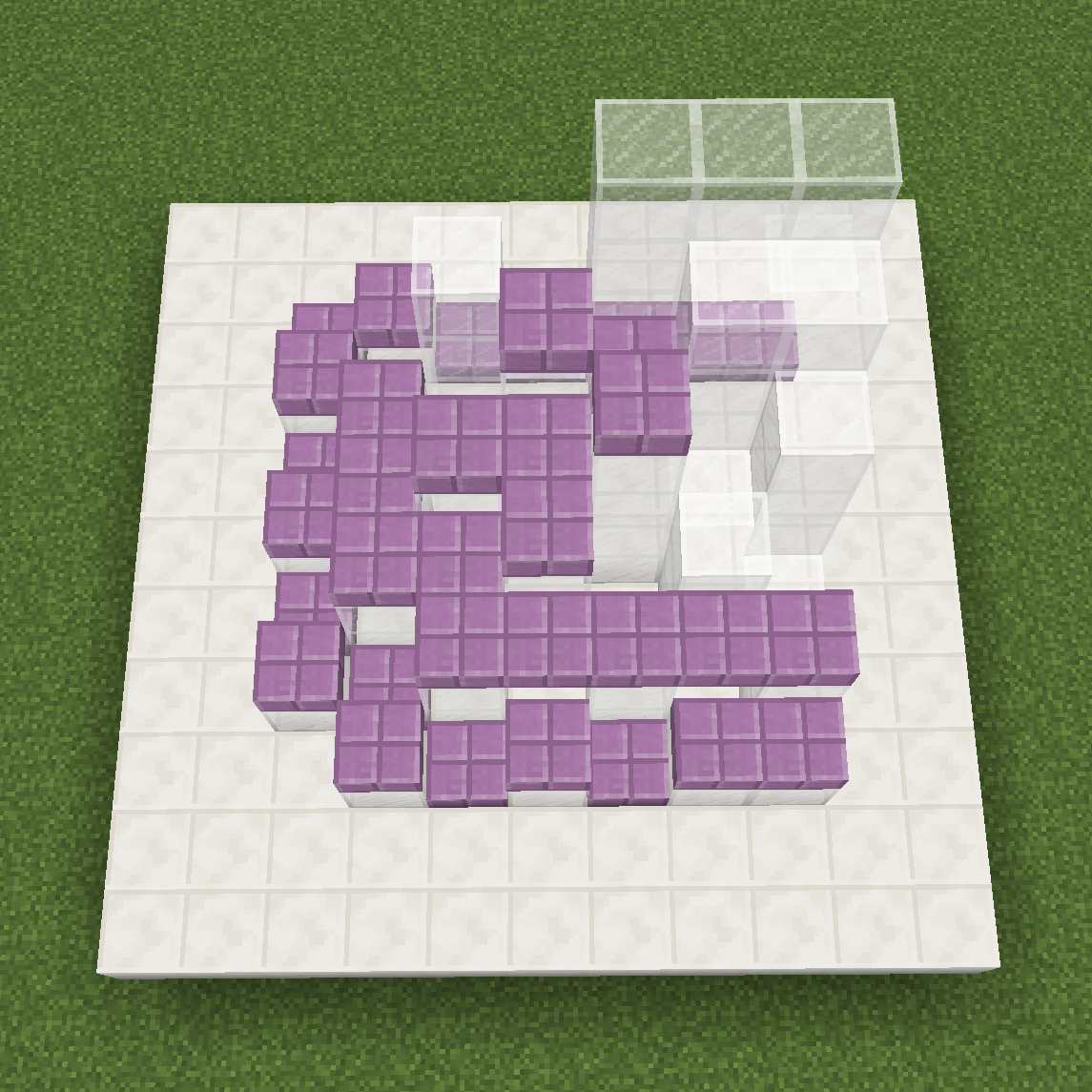}
    \includegraphics[trim={0 0 0 0},clip,width=.49\linewidth]{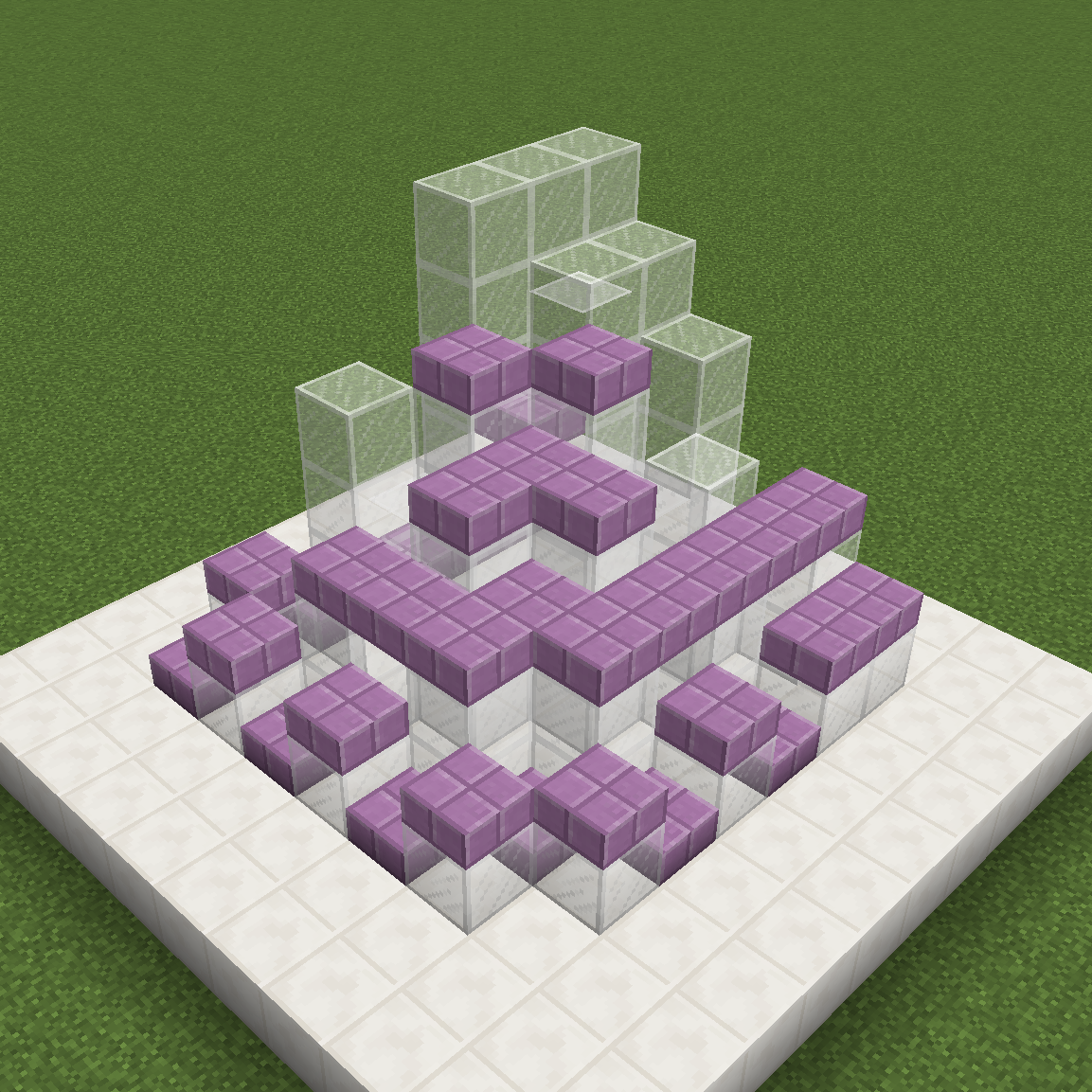}
    \caption{Output of the NCA with highest mean path-length. A dense staircase is generated, spiraling upward in the center of the maze.}
    \label{fig:nca_empt_path_example} 
    \end{subfigure}
    \begin{subfigure}[t]{.49\linewidth}
    \includegraphics[draft=false,width=1\linewidth]{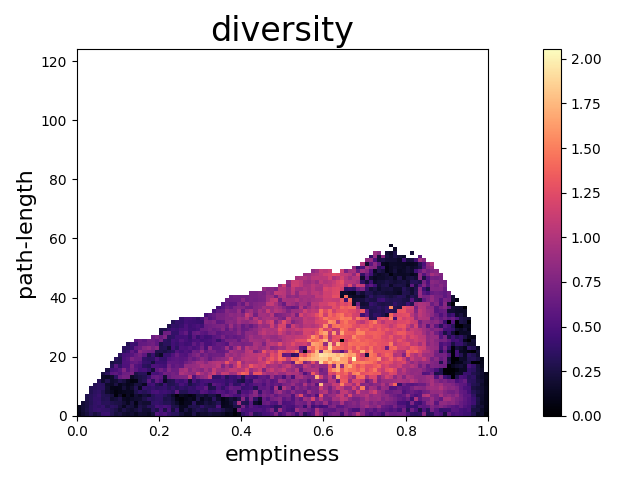}
    \caption{Generators producing levels with medium emptiness and path-length exhibit greater diversity}
    \label{fig:nca_empt_path_diversity}
    \end{subfigure}
    \begin{subfigure}[t]{.49\linewidth}
    \includegraphics[draft=false,width=1\linewidth]{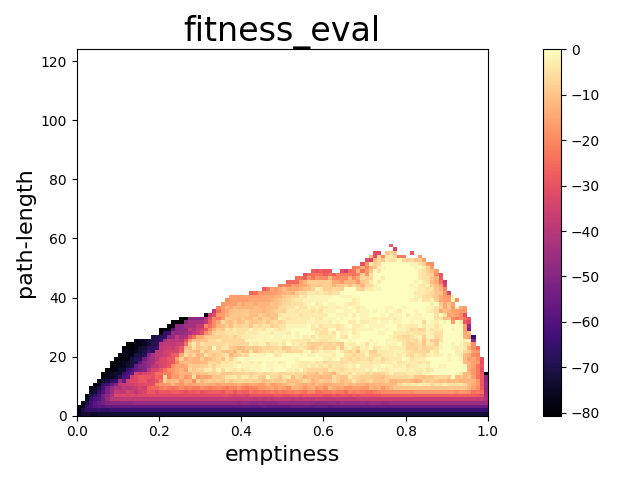}
    \caption{Generators with low path-length and/or emptiness are less fit (with fewer jumps).}
    \label{fig:nca_empt_path_fitness}
    \end{subfigure}
    \caption{Illuminating diverse specialist NCAs to generate mazes with variable diameter and proportion of empty tiles.}
    \label{fig:nca_empt_path}
\end{figure}

\subsection{Evolving diverse generators}
In Fig.~\ref{fig:nca_empt_path}, we show the result of using quality diversity to illuminate diverse NCA level-generators to produce mazes in the diameter problem.
Similar to the RL generator in Fig.~\ref{fig:diam_final}, the agent builds a central spiral staircase (Fig.~\ref{fig:nca_empt_path_example}.
It makes heavy use of redundant sawtooth-like stairs, particularly in hugging the borders of the maze on the ground floor.

With emptiness and path-length as diversity measures, the generator producing the longest mean path-length/diameter achieves only half of the estimated upper bound of this metric.
The mazes with greatest diameter tend to be mostly empty.
Unlike in 2D domains, where mazes with maximum diameter tend to be half-empty~\cite{earle2021learning}, the increased emptiness in large-diameter mazes is aligns with the fact that the player-agent would need to occupy 2 instead of 1 empty tiles.

The NCAs producing the most diverse levels are found in the ``middle'' of the population, with medium emptiness and path-length (Fig.~\ref{fig:nca_empt_path_diversity}).
As either measure takes on more extreme values, the space of possible levels shrinks.
Meanwhile, generator fitness---which is dominated by the levels' containing the target number of jumps---suffers notably when path-length is very low, and when the level has little empty room but relatively high path-length (Fig.~\ref{fig:nca_empt_path_fitness}).
The former results from the fact that to include several jumps, a path must be of non-trivial length (the cumulative length of these jumps).
The latter results from inclusion of a large amount of head-space and otherwise empty tiles in each jump.

\section{Discussion}

In each of the tasks, agents are incentivized to maximize path-length (or diameter).
Often, the agent constructs a large number of stairs, even when these appear ``redundant'' (i.e. descending immediately after ascending, as in Fig.~\ref{fig:nca_empt_path}).
This focus on stairs is understandable, given that each stair-climbing action corresponds to a path-length of three (the upper step, the lower step, and tile above the lower step are all considered as being traversed in our path-finding algorithm).
This corresponds to high path-length ``value'' relative to space required and the complexity of the build.
While jump actions (between tiles one apart in any of the four cardinal directions, separated by a chasm, and with sufficient headroom above them) similarly correspond to a path-length of 3, there are more constraints on what constitutes a valid jump, and more space is required to build a jump.

We note that in many of the tasks we set out for the RL-based generators, a large recurring structure (e.g. a climbable central mountain) or particular item placement (e.g. a chest in the same corner of a dungeon) reoccurs across multiple episodes. There are often minimal modifications to this underlying structure, making the generated level quickly amenable to the particular control task in the episode at hand (i.e. placement of doors or target number of jumps/path-length).

This behavior is understandable given that, unlike what was done in the original PCGRL paper, we do not limit the degree to which the agent may change the initial board state~\cite{khalifa2020pcgrl}. Instead we emphasize generative diversity as induced by potentially user-controllable aspects of the environment.
While it is always possible to make more aspects of the level (like the placement of the chest) controllable, future work might seek to produce more general content diversity even beyond these user controls, for example by limiting the degree to which the agent changes a random initial state, or by having a second generator-agent explore the space of constraints on the primary generator (e.g. by ``locking'' certain tiles such as a chest in place) so as to maximize the latter's learning progress.

In evolving diverse populations of NCA generators, we appear to overshoot the boundaries of the path-length and number-of-jump metrics.
Whereas in 2D environments, it is relatively simple to construct a maximum path-length maze on paper, and use this to inform our upper bound, this becomes more difficult in 3D dimensions, with stairs and jumps included in our path-finding logic.

Future work could replace the rough estimate on the upper bound of path-length used here with something tighter, preferably by describing algorithms that generate paths with maximum length or number of jumps.
Since scaling to more complex domains might make such manual solutions infeasible, another approach could be to use a quality diversity method in which upper/lower bounds on the possible values taken on by diversity measures are not required in advance (e.g. Sliding Bounds MAP-Elites~\cite{fontaine2019mapping}).

It is often argued that PCG in games can alleviate the authorial burden by doing some content generation autonomously or in combination with a human designer. While acknowledging that the ability of a PCG solution to actually help designers with their work is highly dependent on the ways in which the system can be controlled and interacted with by the designer, it is also worth highlighting that some types of content require more human effort to create than others. Arguably, three-dimensional content needs more authorial effort, partly because of the complexities outlined above and partly because there's simply more of it, and thus more interesting to develop PCG solutions for. 

Controllable PCGRL, and the evolution of diverse generators for MDP-style level-design tasks, are both suitable candidates for future use in mixed-initiative PCG systems.
This is partly because, compared to SBPCG, PCGRL front-loads the training time; once a model is trained (which can take long time), new content can be generated almost instantaneously. The other reason is that RL-trained generators are iterative, making one change at a time, which allows for easier turn-taking between human and machine~\cite{delarosa2021mixed}.

Another downstream application of the generators produced here could be in generating curricula of environments of learning (player) agents, as in Unsupervised Environment Design (UED)~\cite{bontrager2021learning, dennis2020emergent, jiang2021replay, parker2022evolving}.
Where these methods for co-learning agents and environments may be limited in terms of the representational capacity of their environment generators, our (populations of) generators could be pre-trained to produce sets of levels with controllable low- and high-level features.
During player training, a ``curator'' agent could act on these controls in response to player learning progress.
This could allow for a curriculum with more fluid and efficient variation.

Conversely, sets of MDP-based level generators could be co-learned to maximize each others' regret, learning progress, or otherwise.
Rather than training generators to merely respond to moving doors, for example, they could be trained to work around arbitrary subsections of the level fixed in place by another, ``curatorial'' generator.
Such work would benefit from the co-learning incentives introduced by UED, which ensure that tasks remain at an appropriate level of difficulty.

In experimenting with RL and evolutionary methods for MDP-based PCG in parallel, we note that it could be fruitful to combine these methods, training a ``generalist,'' controllable generator by imitating an evolved population of specialist models.
\section{Conclusion}

In this work, we apply Procedural Content Generation via Reinforcement Learning and Quality Diversity to 3D game environments, training generators to produce maximally complex mazes and dungeons between user-defined entrance and exit positions. We introduce several 3D PCG tasks which could be used to evaluate the capabilities of iterative generators. These tasks contain elements that highlight 3D domain affordances, such as a third dimension, jumping, and gravity.
We find that the generator adapted from the 2D PCGRL framework~\cite{khalifa2020pcgrl} is able to satisfy functional constraints and maximize level complexity while working around fixed structures (i.e. entrances and exits).


One challenge with three-dimensional content that we do not address in this paper is that the atomic content units are often not uniform in shape and size. Whereas 2D games are typically based on tiles or similar uniform units, many 3D games have assets of highly varying shape as primitives. We sidestep this issue by focusing on Minecraft, which uses voxels. Future tasks could include non-voxel-based environments to further push the boundaries of 3D PCGRL.

\bibliographystyle{ACM-Reference-Format}
\bibliography{references}

\end{document}